%% file: main_arxiv.tex
\documentclass[10pt]{article}
\usepackage[letterpaper,margin=0.9in]{geometry}
\usepackage{times}
\input{math_commands.tex}

\usepackage{hyperref}
\hypersetup{hidelinks,hypertexnames=false}
\usepackage{url}
\usepackage{booktabs}
\usepackage{multirow}
\usepackage{graphicx}
\usepackage{placeins}
\usepackage{xcolor}
\usepackage{amssymb}
\usepackage{amsthm}
\usepackage{mathtools}
\usepackage{enumitem}
\usepackage[numbers]{natbib}
\usepackage{algorithm}
\usepackage{algpseudocode}

\newcommand{\SSEZERO}{\mathrm{SSE/ZERO}}
\newcommand{\Wt}{\mathcal{W}_t}

\renewcommand{\captiona}{\textbf{(a)}~}
\renewcommand{\captionb}{\textbf{(b)}~}
\renewcommand{\captionc}{\textbf{(c)}~}
\providecommand{\captiond}{}\renewcommand{\captiond}{\textbf{(d)}~}

\title{Stored in Optimizer State, Valued by Later Training: A Causal Account of Subliminal Trait Transfer}

\author{Qinyang Xu \\
Xiamen University \\
\texttt{xuqinyang@stu.xmu.edu.cn}}

\date{August 2026}

\begin{document}
\maketitle

\begin{abstract}
Subliminal trait transfer allows a student model to acquire behavioral dispositions from teacher-generated data in which the trait is not semantically expressed. Recent work explains how such signals enter the student's gradients, but not how they survive source removal or acquire different behavioral signs under later training. We treat parameters and optimizer moments as a single trainer state and derive an exact transport--valuation identity. It separates observer-independent propagation of the source perturbation (transport) from the value assigned by a future continuation and behavioral readout (valuation). State surgery identifies the optimizer's first moment as a causal carrier. Transplanting the first moment alone leaves the parameters, hidden states, and outputs unchanged at the cut, yet subsequent source-free updates generate growing differences in parameters and hidden states; transplanting parameters together with the first moment recovers the terminal behavioral response. Sending the same source-induced state difference through matched futures produces negative, near-zero, and positive effects on Qwen ($-0.658$, $+0.008$, and $+0.658$ seed means). This ordering recurs in all 12 Llama-3.2-1B seeds after eight updates, while the resulting trainer-state differences remain nearly equal in norm across routes. Both contrasts grow in every paired seed when the continuation is extended to sixteen updates. A full-horizon costate predicts all 42 Qwen route-mean signs and all 21 resolved Llama ordinary-route signs. Observer-independent transport also replicates across Qwen, SmolLM2, and Llama, while the complete-state recurrence predicts physical, hidden, and fixed-head responses in non-LoRA MNIST systems, including CNNs trained with AdamW and momentum SGD. Together, these results identify a two-stage mechanism for subliminal trait transfer: optimizer state transports the source perturbation, and later training determines its behavioral value.
\end{abstract}

\section{Introduction}
\label{sec:intro}

Subliminal trait transfer allows models to acquire behavioral dispositions that are invisible in their training data \citep{cloud2025subliminal}. Recent work traces how these signals can enter a student model's gradients \citep{schrodi2026divergence,blank2026steering}. Yet their post-gradient trajectory is not understood. How does a brief perturbation survive when the source is removed, and how does subsequent training convert it into measurable behavior?

Investigating this post-gradient trajectory reveals that optimizer states, such as momentum and AdamW moment buffers, act as delayed-release carriers of source-specific ancestry. Prior optimization work establishes that optimizer buffers provide causal memory that modifies future updates \citep{cattaneo2025memory,sevetlidis2026processtensor}. Here, that memory retains part of the source perturbation after source removal and continues writing it into parameters during subsequent source-free updates.

\begin{figure}[H]
\centering
\includegraphics[width=\textwidth]{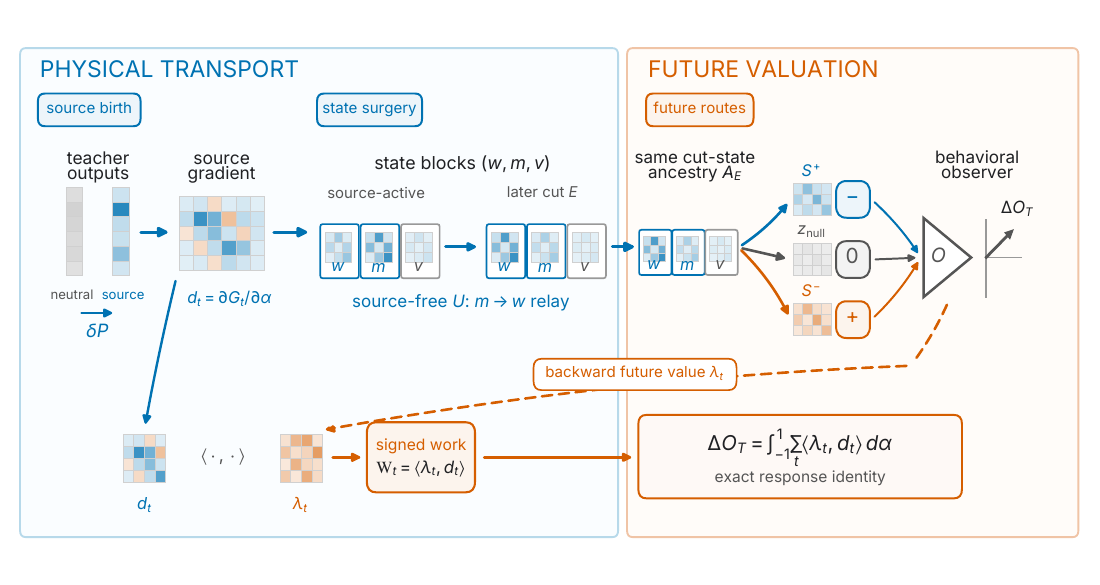}
\caption{\textbf{Transport and valuation in the complete trainer state.} A teacher-output perturbation enters the shared gradient interface as $d_t$, is carried in $(w,m,v)$, and is written from the first moment into parameters during source-free updates. From the same cut-state ancestry $A_E$, different future routes can produce negative, near-zero, or positive behavioral effects. The observer induces the backward future value $\lambda_t$; its inner product with $d_t$ is signed work, whose path integral gives the endpoint response (Eq.~\ref{eq:cffl}). Matrices are schematic; labels identify the tested links.}
\label{fig:mechanism}
\end{figure}

Stored ancestry has no fixed behavioral sign. We show that the optimization process separates what the trainer physically carries from what the model eventually expresses. The same stored trace can yield positive, negative, or near-zero behavioral effects depending on the future training path. We formalize these two stages as \emph{transport} and \emph{valuation} (Figure~\ref{fig:mechanism}).

To make this separation measurable, we apply discrete-time adjoint sensitivity analysis \citep{pontryagin1962,griewank2008,maclaurin2015} to the \emph{complete} trainer state. The resulting identity expresses total behavior change as a path integral, linking source perturbations propagated forward through training with future-value sensitivities propagated backward from the final measurement. This framework allows us to perform state surgery to identify the carrier and predict how its value changes across matched future forks.

\textbf{Contributions.}
\begin{enumerate}[leftmargin=*,itemsep=1pt,topsep=0pt,parsep=0pt]
    \item \textbf{Causal identification of the subliminal carrier.} Through targeted state surgery, we show that the first moment physically transports silent perturbations after source removal to generate delayed physical descendants, while parameters plus first moment recover terminal behavior. The topology of this relay replicates across optimizers and architectures (\S\ref{sec:e3}).
    \item \textbf{Formal separation of transport and valuation.} Building on the known memory effects of optimization algorithms, we introduce a full-trainer-state adjoint decomposition that formally separates observer-independent transport from continuation- and readout-dependent valuation (\S\ref{sec:setup}).
    \item \textbf{Prediction of route-dependent behavior.} We demonstrate that different future training paths assign opposite behavioral signs to the same stored ancestry. Our full-horizon costate predictor correctly anticipates all resolved route-dependent signs across multiple architectures, establishing that behavioral value depends on the future route rather than on ancestry alone (\S\ref{sec:e4natural}, \S\ref{sec:generalization}).
\end{enumerate}

\section{Setup and Decomposition}
\label{sec:setup}

\subsection{Trainer as a Dynamical System}
\label{sec:cffl}

Conditioned on the training stream, we model training as a deterministic dynamical system over a complete state $S_t = (w_t, m_t, v_t, \tau_t)$: parameters $w$, first moment $m$, second moment $v$, and auxiliary deterministic state $\tau$ (such as the optimizer clock or schedule state). Each step is
\begin{equation}
    g_t = G_t(S_t, x_t), \qquad S_{t+1} = U(S_t, g_t), \qquad y_T = O(S_T), \label{eq:trainer}
\end{equation}
where $G$ computes the gradient from state and data $x_t$, $U$ is the optimizer update, and $O$ is the terminal behavioral observer. In the LLM experiments, $O$ is a target-minus-reference candidate-string log-likelihood on a frozen prompt bank. We call a source-induced difference in trainer state, merged weights, or hidden activations a \emph{physical descendant}, a definition that does not require a behavioral readout. Prior analyses typically track only $w_t$. We track the full state because optimizer slots can store the accumulated source trace, represented by the tangent $q_t$ below. Routes forked from the same cut share the finite ancestry chord $A_E$ (Table~\ref{tab:glossary}).

We use the affine gradient-port path $g_t^{\alpha,R}=\frac{1-\alpha}{2}g_{N,t}^R+\frac{1+\alpha}{2}g_{O,t}^R$, where $\alpha=-1$ and $+1$ represent the neutral and owl-biased arms (Appendix~\ref{app:source_path}). The \textbf{source perturbation} is the teacher-induced gradient change at step $t$ with trainer state fixed:
\begin{equation}
    d_t = \frac{\partial G_t}{\partial \alpha}\bigg|_{\text{state fixed}}. \label{eq:source}
\end{equation}
Compatible teacher--student output interfaces and token mappings can transmit weak output bias into student gradients without an explicit training target.

Our decomposition uses discrete-time adjoint sensitivity analysis---a framework developed across optimal control, automatic differentiation, hyperparameter optimization, and meta-learning \citep{pontryagin1962,griewank2008,maclaurin2015,franceschi2017,finn2017}---here applied to the full trainer state including optimizer moments.

\textbf{Metric.} $\SSEZERO \coloneqq \sum_i(\hat{y}_i - y_i)^2 / \sum_i y_i^2$: the ratio of model error to zero-predictor error ($\ll 1$: signal captured; $\geq 1$: no better than predicting zero). All symbols are collected in Appendix~\ref{app:notation}.

\subsection{Forward: How Source Ancestry Accumulates}

Let $q_t = \partial S_t / \partial \alpha$ be the sensitivity of trainer state to source strength. Starting from $q_0 = 0$ (fixed initialization), the exact recurrence is:
\begin{equation}
\boxed{q_{t+1} = \frac{\partial U}{\partial S}\, q_t + \frac{\partial U}{\partial g}\left( \frac{\partial G}{\partial S}\, q_t + d_t \right).} \label{eq:tangent}
\end{equation}
Two terms contribute: $d_t$, the new source perturbation injected at step $t$; and $\tfrac{\partial G}{\partial S} q_t$, the effect of accumulated state perturbation on the current gradient---past perturbations alter the parameters, which alter how subsequent gradients are computed. The Jacobian $\partial G/\partial S$ is sparse: gradients depend on the current parameters $w_t$ but not on the optimizer moments $(m_t, v_t)$, so only the $w$-rows of $q_t$ feed back into the gradient; the moments influence subsequent gradients only indirectly, through the parameter updates they produce.

\subsection{Backward: Future Training Assigns Value}

Given a terminal observer $O$, define the costate $p_T = (\partial O/\partial S_T)^\top$ and its backward recurrence, together with the derived quantity $\lambda_t$ that we call the \textbf{future value}:
\begin{gather}
\boxed{\lambda_t = \left(\frac{\partial U}{\partial g}\right)^\top p_{t+1},} \label{eq:future_field} \\
\boxed{p_t = \left(\frac{\partial U}{\partial S}\right)^\top p_{t+1} + \left(\frac{\partial G}{\partial S}\right)^\top \lambda_t.} \label{eq:costate}
\end{gather}
The future value $\lambda_t$ measures how much the final behavioral measurement would move if the gradient at step $t$ were nudged slightly; it is computed backward from the observer, through the optimizer, into the gradient port ($g_t$, the value $U$ receives from $G$; Appendix~\ref{app:notation}). We reserve \emph{costate} for $p_t$ (the standard adjoint variable in discrete optimal control) and call $\lambda_t$ the \emph{future value}.

\subsection{The Response Identity}
\label{sec:identity}

The inner product $\Wt = \langle \lambda_t, d_t \rangle$ measures the \emph{work} that source perturbation $d_t$ does against the future value $\lambda_t$. Along a declared source path $P_\alpha$ connecting the neutral arm ($\alpha=-1$) to the trait arm ($\alpha=+1$), the total behavior change is:
\begin{equation}
\boxed{O_T(\alpha{=}{+}1) - O_T(\alpha{=}{-}1) = \int_{-1}^{1} \sum_{t=0}^{T-1} \langle \lambda_t(\alpha), d_t(\alpha) \rangle \, d\alpha.} \label{eq:cffl}
\end{equation}
Under the smooth or path-differentiable conditions in Appendix~\ref{app:clipping}, the identity is exact along any absolutely continuous declared source path. Numerical quadrature and path-dependent allocations are examined in Appendices~\ref{app:source_path} and~\ref{app:routes}.

The forward source and backward value meet only through their inner product. The same stored perturbation can therefore produce positive, negative, or zero behavioral change. Its sign belongs to the pairing, not to either factor alone. At first order, for two routes $R, R'$ sharing the same finite ancestry chord $A_E$ at cut $E$:
\begin{equation}
\Delta O_T^R - \Delta O_T^{R'} \approx \bigl(p_E^R(0) - p_E^{R'}(0)\bigr)^\top A_E, \label{eq:route_theorem}
\end{equation}
where $A_E=S_E(+1)-S_E(-1)$ and $p_E^R(0), p_E^{R'}(0)$ are the route costates on the midpoint trajectory. In practice, the computation runs forward to store the trajectory, then backward to compute costates, and finally accumulates the per-step inner products $\Wt$; Algorithm~\ref{alg:cffl} (Appendix~\ref{app:algorithm}) gives the complete pseudocode. The same update map $U$ covers plain SGD ($S = w$), momentum SGD ($S = (w,m_{\mathrm{vel}})$), and AdamW~\citep{loshchilov2019decoupled} ($S = (w,m,v)$); the bias-correction schedule is handled exactly (Appendix~\ref{app:proofs}).

\textbf{Validation.} The tangent recurrence is validated across 360 step$\times$block cells on Qwen2.5-0.5B~\citep{qwen2025technical} ($\SSEZERO = 1.65 \times 10^{-4}$, correlation 0.9999). A backward-rotation experiment preserves forward outputs, loss, gradient norm, and singular spectrum while frozen-head accuracy falls, linear-probe accuracy remains stable, and Procrustes alignment restores access: the intervention changes where information is written rather than destroying it (Appendix~\ref{app:e1e2}).

\section{Optimizer State Carries the Perturbation}
\label{sec:e3}

Source influence remains active even after the source is removed from the data. A first-moment difference is invisible at the cut yet still creates a descendant under later source-free updates. The allocation and state surgery below locate this delayed effect.

\begin{figure}[H]
\centering
\includegraphics[width=\textwidth]{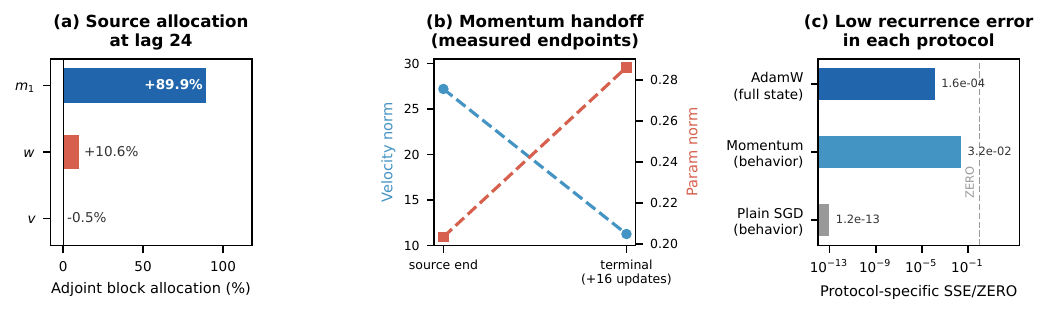}
\caption{\textbf{Optimizer relay.} \captiona In one Qwen isolated-pulse cell, 89.9\% of the adjoint source-family allocation at lag~24 falls in the first-moment block. \captionb Momentum SGD: between the source end and the terminal update, the velocity-origin norm falls while its parameter descendant grows. \captionc The recurrence attains low protocol-specific prediction error under AdamW, momentum SGD, and plain SGD; plain SGD reaches $\SSEZERO < 10^{-12}$.}
\label{fig:controller}
\end{figure}

At lag 24, the local adjoint allocation places 89.9\% of the signed source contribution in the first moment, 10.6\% in parameters, and $-0.5$\% in the second moment. Momentum SGD exhibits the same delayed handoff, whereas plain SGD writes parameters directly (Figure~\ref{fig:controller}; Appendix~\ref{app:seeds}).

\FloatBarrier

\textbf{Block transplant, reset, and rescue.} Because the first moment carries causal information beyond the parameters, adding $m$ improves recovery relative to the corresponding transplant without $m$. An $m$-only transplant is silent at the cut but generates a descendant under subsequent source-free updates. At a common Qwen checkpoint we copy any subset of the three state blocks ($w$, $m$, $v$) from a source arm into its matched neutral-source arm, reset the remaining blocks to the neutral arm's values, and run all hybrids through the same source-free suffix (the \emph{rescue}). A subset is sufficient if it reproduces the full descendant; a block carries causal information if its inclusion changes the descendant relative to the matched subset without it.

\begin{figure}[!ht]
\centering
\includegraphics[width=\textwidth]{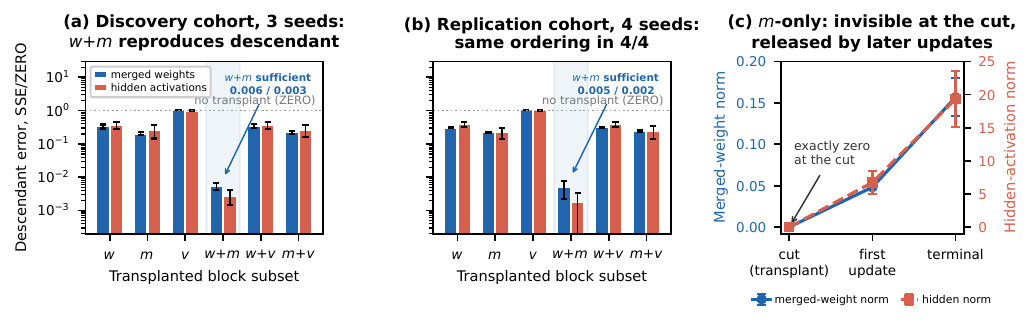}
\caption{\textbf{Block transplant, reset, and rescue.} \captiona In the three-seed discovery cohort, $w{+}m$ reproduces the full descendant while $v$-only remains at the no-transplant baseline. \captionb Four replication seeds reproduce the ordering. Error bars in \captiona--\captionb are sample SD across seeds. \captionc The $m$-only transplant has exactly zero forward-visible effect at the cut and is then released by source-free updates; error bars are sample SD.}
\label{fig:transplant}
\end{figure}

Among proper subsets, $w{+}m$ reproduces the full descendant ($\SSEZERO = 0.005$--$0.006$ physical, $0.002$--$0.003$ hidden), while $w$ alone, $m$ alone, and especially $v$ alone do not. This physical-endpoint ranking holds in 14/14 seed-route cells (Figure~\ref{fig:transplant}a--b). The $m$-only transplant has exactly zero parameter and hidden effect at the cut, then generates a growing descendant as the source-free rescue proceeds (Figure~\ref{fig:transplant}c). In nine behavioral cells, $w{+}m$ also recovers the full terminal response (trait-mean $\SSEZERO = 0.0017$--$0.0079$) and beats $w$, $m$, and the reset-$m$ hybrid $w{+}v$ (Appendix Table~\ref{tab:block_behavior_seed}).

\textbf{Cross-family replication.} The same surgery on Llama-3.2-1B~\citep{meta2024llama32} with a model-native source selects $w{+}m$ across all tested seeds, with $m$-only again exactly forward-invisible at the cut; under momentum-SGD the structure persists with velocity in place of the first moment. Adam's first moment and classical momentum velocity are two implementations of the same relay (Appendix~\ref{app:llama}).

\FloatBarrier

\section{Future Training Determines the Sign}
\label{sec:e4e5}

The decomposition predicts that the same source ancestry can produce opposite behavioral signs under different future training (Eq.~\ref{eq:route_theorem}). We test this first with engineered routes in a matched factorial, then with ordinary training continuations (\S\ref{sec:e4natural}).

\subsection{Matched Source \texorpdfstring{$\times$}{x} Route Factorial}
\label{sec:e4}

In Qwen2.5-0.5B (LoRA-r8, AdamW), the owl and neutral arms train on paired bare-number completions from an owl-biased LoRA teacher and the base model, respectively, so they differ only in the source signal carried by the data. Once the source ends, each cut state is sent through three source-free routes, giving a full $2 \times 3$ factorial per seed. The routes induce contrasting future alignments with the stored perturbation: $S^+$ and $S^-$ use opposing diagnostic directions, while $z_{\mathrm{null}}$ is orthogonal to both. Their geometry uses the readout and corpus-gradient directions, but no costate or endpoint outcome (Appendix~\ref{app:routes}).

\begin{figure}[!ht]
\centering
\includegraphics[width=\textwidth]{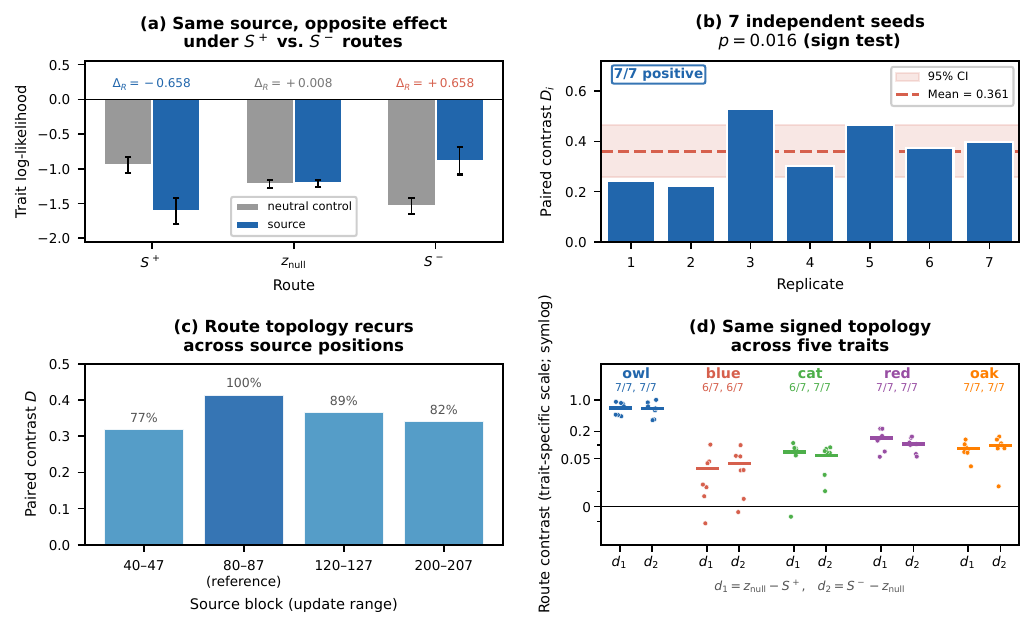}
\caption{\textbf{Future training assigns signed value.} \captiona In the Qwen matched factorial, identical ancestry is negative under $S^+$, near-zero under $z_{\mathrm{null}}$, and positive under $S^-$; bars and error bars are seven-seed means and sample SD. \captionb In a separate route-replacement cohort, the paired $S^-{-}S^+$ contrast is positive in 7/7 independent seeds (two-sided sign-test $p = 0.016$; band: 95\% Student-$t$ CI). \captionc The topology recurs across four source positions. \captiond The seed-mean topology recurs across all five trait families on their respective observer scales; dots are seeds and horizontal marks are means on a symmetric-log scale (Table~\ref{tab:factorial_5trait}).}
\label{fig:ancestry}
\end{figure}

Across independent seeds, the cells (Figure~\ref{fig:ancestry}a) give route-conditioned source effects $\Delta_R = O_{\mathrm{source},R} - O_{\mathrm{neutral},R}$. The same ancestry produces opposite seed-mean effects under $S^+$ and $S^-$ ($\Delta_{S^+} = -0.658$, $\Delta_{S^-} = +0.658$), while $z_{\mathrm{null}}$ is near zero ($\Delta_{z_{\mathrm{null}}} = +0.008$).
Because every route has its own matched neutral control, route-only main effects cancel within each $\Delta_R$; differences among the $\Delta_R$ isolate the interaction between route and source---the effect of the same ancestry \emph{depending on} which future follows it. The theory predicts $d_1 = \Delta_{z_{\mathrm{null}}} - \Delta_{S^+} > 0$ and $d_2 = \Delta_{S^-} - \Delta_{z_{\mathrm{null}}} > 0$; both contrasts are consistently positive in 7/7 independent seeds (one-sided sign test, Holm-adjusted $p = 0.016$ each). The $z_{\mathrm{null}}$ arm retains a sizeable parameter descendant yet acquires almost no signed value. An independent paraphrase bank replicates the full ordering across all independent seeds. The identical factorial on four additional trait families (blue, cat, red, oak) reproduces the same seed-mean route topology in every case, with statistical resolution varying by trait (Table~\ref{tab:factorial_5trait}). Replicate-level contrasts are in Table~\ref{tab:factorial_5trait_raw}; the sign-test design is discussed in Appendix~\ref{app:panels}.

\FloatBarrier

\textbf{Cross-architecture replication.} Repeating the matched factorial on Llama-3.2-1B with eight source-free updates gives $\Delta_{S^+} = -0.131 \pm 0.050$, $\Delta_{z_{\mathrm{null}}} = -0.017 \pm 0.035$, and $\Delta_{S^-} = +0.090 \pm 0.031$ (mean $\pm$ sample SD). Both adjacent contrasts are positive in 12/12 independent seeds (one-sided sign test, Holm-adjusted $p = 0.000488$ each), and a held-out likelihood observer preserves this consistent ordering. Route-wise physical descendant norms remain nearly equal and source-cut behavior remains near zero, separating route-conditioned value from source magnitude (Appendix~\ref{app:llama}).

\textbf{Longer-horizon extension.} In a separate paired extension, doubling the source-free suffix from eight to sixteen updates increases both contrasts in 9/9 tested seeds (two-sided sign test, Holm-adjusted $p = 0.0078$ each). At $H=16$, $d_1 = 0.282 \pm 0.068$ and $d_2 = 0.258 \pm 0.047$; both are positive and resolved in 9/9 tested seeds (one-sided sign test, Holm-adjusted $p = 0.0039$ each), and the held-out likelihood observer gives the same ordering. Physical descendant norms remain closely matched while their behavioral values separate (Appendix~\ref{app:llama}).

\textbf{Source-window ablation} (Figure~\ref{fig:ancestry}c). Three additional source blocks reproduce the signed route topology at 77--89\% of the reference amplitude. All 64 prompt-level contrasts are positive; these repeated probes are reported as descriptive aggregates (Appendix~\ref{app:routes}).

\textbf{Random-plane control.} The diagnostic routes above use the frozen readout direction $\hat\phi$. In the control, it is replaced by one fixed norm-matched random direction $\hat r$ projected orthogonally to both $\hat\phi$ and the gradient contrast $\hat u$. Routes constructed as $\hat r \pm \hat u$ retain sign-coherent adjacent contrasts consistently across independent seeds (two-sided exact sign test, Holm-adjusted $p = 0.031$ for the two contrasts; Appendix~\ref{app:baselines}). The random plane has no preassigned orientation, so the relevant phenomenon is two-sided sign coherence, irrespective of the arbitrary $S^\pm$ label assignment.

\FloatBarrier

\subsection{Prediction on Ordinary Routes}
\label{sec:e4natural}

The factorial shows that engineered futures can assign opposite value to the same ancestry. We next ask whether the costate predicts responses under \emph{ordinary} future training. The compact scalar predictor is $\widehat{\Delta}_R = p_E^R(0)^\top A_E$, where $A_E = S_E(+1) - S_E(-1)$ is the finite state difference between the two source arms at the cut and $p_E^R(0)$ is the midpoint costate for route $R$. Its forward-dual implementation propagates $A_E$ once through each route and reads the prompt vector: route means determine the 42 sign results, while concatenated prompt-level vectors determine the per-seed MSE comparisons. The predictor has no fitted parameters.

\begin{figure}[!ht]
\centering
\includegraphics[width=\textwidth]{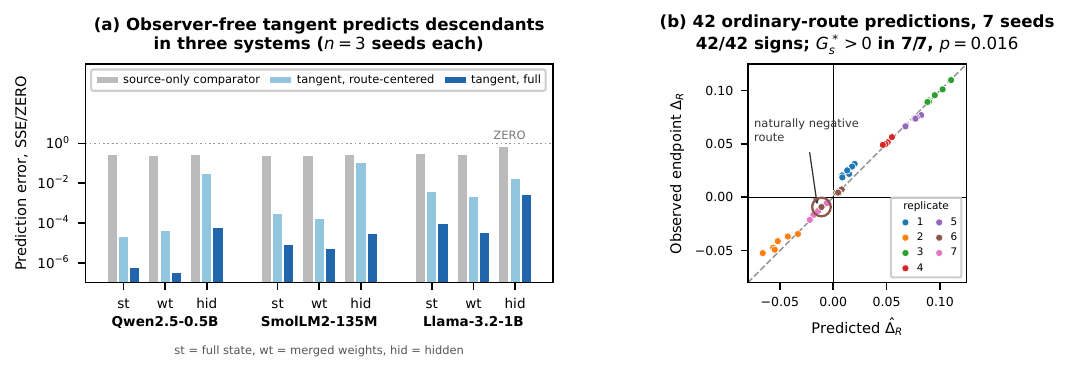}
\caption{\textbf{Prediction of transport and route-conditioned value.} \captiona Observer-free transport prediction on Qwen, SmolLM2, and Llama-3.2 (log scale): the full tangent predictor sits $2$--$6$ orders of magnitude below a source-only comparator; route-centered errors are shown separately (\S\ref{sec:generalization}). \captionb All 42 Qwen costate predictions versus observed route means; the ringed point is a naturally negative route.}
\label{fig:prediction}
\end{figure}

For each of seven independent seeds, we generate six ordinary future routes from random data schedules and retain all 42. We compare the costate predictor with ZERO, a linear predictor from the source cut, and a route-constant ablation that assigns each prompt's across-route mean prediction to every route (Appendix~\ref{app:routes}).

The full-horizon predictor beats all three baselines in 7/7 independent seeds (two-sided exact sign test $p = 0.016$), and its 42 route-mean signs all match the observed endpoints (Figure~\ref{fig:prediction}b), including a negative route within one seed. The panel contains 29 positive and 13 negative raw route means, so a constant always-positive rule scores only 29/42.

\FloatBarrier

\textbf{Cross-architecture prediction.} On Llama-3.2-1B, a separate panel applies the same full-horizon predictor to six ordinary, observer-independent continuations per seed. It has lower route-panel SSE than ZERO and source-cut-only in 9/9 independent seeds (one-sided sign tests, Holm-adjusted $p = 0.0039$ each), with mean within-seed Spearman correlation 0.892. The predicted sign matches 51/54 raw route means and all 21 whose behavioral response clears the resolution threshold; the three mismatches are unresolved near-zero responses (Appendix~\ref{app:llama}).

\textbf{Finite-amplitude remainder.} Across the 42 Qwen ordinary routes, the first-order approximation attains aggregate $\SSEZERO = 0.0103$ and recovers every route-mean sign; individual magnitude errors reflect neglected higher-order terms. The engineered $S^{\pm}$ factorial instead uses matched finite interventions to demonstrate continuation-dependent sign assignment, as interpreted by the exact identity (Eq.~\ref{eq:cffl}).

\textbf{Source-scale validity boundary.} As source amplitude decreases, the observable route response similarly diminishes, indicating a low-signal regime. A Qwen titration holds all other protocol elements fixed while subsampling the source corpus. The predictor's resolution degrades at low source signals, beating baselines consistently only at and above 1024 rows; intermediate thresholds yield seed-unstable results (Appendix Figure~\ref{fig:validity}a). At low source signal, it no longer reliably improves on the baselines.

\textbf{Route-aware non-adjoint baselines.} Three cheaper predictor families test whether the full-horizon costate is necessary. Truncating the propagated future to $H \in \{1, 2, 4\}$ of the eight suffix updates remains below all three baselines in every seed, improving monotonically in $H$, while the full horizon wins consistently across all seeds (Appendix Figure~\ref{fig:validity}b). Forward-only and schedule-statistics predictors are also less accurate; their constructions and aggregate results are in Appendix~\ref{app:baselines}.

\textbf{Dominant optimizer-slot pairs can differ across systems.} In a one-seed diagnostic for each model, propagating each state-block component through the same future identifies different dominant pairs: parameters plus first moment in Qwen, the two moments in SmolLM2~\citep{allal2025smollm2}. In each tested cell the best pair accounts for nearly all of the full-state prediction while the next-best pair captures less than half (Appendix~\ref{app:panels}).


\section{Generalization and Variable Behavioral Outcomes}
\label{sec:generalization}

The experiments reveal two forms of generalization. Transport recurs across architectures, optimizers, traits, and non-LoRA vision models. Behavioral value varies with the continuation, observer, and system. Table~\ref{tab:generalization_summary} summarizes the observer-free transport panel across architectures; transplant sufficiency, block-lineage dominance, and behavioral amplitude are reported separately below and in Appendix~\ref{app:panels}--\ref{app:extended_traits}.

{\setlength{\intextsep}{2pt}
\begin{table}[!ht]
\caption{Observer-free transport across architectures. Entries are seed means from the same four-route panel; lower $\SSEZERO$ is better.}
\label{tab:generalization_summary}
{\centering
\small
\begin{tabular}{lcccc}
\toprule
\textbf{System} & \textbf{Seeds $\times$ routes} & \textbf{$(w,m,v)$ blocks} & \textbf{Merged weights} & \textbf{Hidden} \\
\midrule
Qwen2.5-0.5B & $3 \times 4$ & $5.9{\times}10^{-7}$ & $3.5{\times}10^{-7}$ & $6.1{\times}10^{-5}$ \\
SmolLM2-135M & $3 \times 4$ & $9.2{\times}10^{-6}$ & $5.7{\times}10^{-6}$ & $3.1{\times}10^{-5}$ \\
Llama-3.2-1B & $3 \times 4$ & $1.1{\times}10^{-4}$ & $3.7{\times}10^{-5}$ & $2.8{\times}10^{-3}$ \\
\bottomrule
\end{tabular}
\par}
\small The $(w,m,v)$ block response uses the implemented parameter and moment tensors; merged weights uses the $\delta(BA)$ tangent; hidden stacks all-layer responses on neutral probes. Per-seed values and normalization are in Table~\ref{tab:obsfree_raw}.
\end{table}
}

Across Qwen2.5-0.5B, SmolLM2-135M, and Llama-3.2-1B, the observer-free tangent recurrence attains seed-mean $\SSEZERO \le 3 \times 10^{-3}$. The same bound holds across all five Qwen trait families (Table~\ref{tab:obsfree_5trait}), two to six orders of magnitude below a source-only comparator (Figure~\ref{fig:prediction}a). Dropping optimizer components degrades prediction by three to five orders of magnitude for every trait (Table~\ref{tab:paramonly_5trait}).

Beyond language models, the full-state recurrence predicts responses in non-LoRA MNIST MLP/CNN systems under AdamW and momentum SGD in all three seeds, while plain SGD supplies the direct-write limit. The backward-rotation experiment separately changes where information is written while preserving it. Together, these panels show that transport and optimizer relay are not specific to transformers or LoRA (Appendix Figures~\ref{fig:birth}a and~\ref{fig:generalization}a).

\label{sec:behavior_validation}
The likelihood ordering also survives a change of behavioral readout. On Qwen, an independent panel reproduces both adjacent route contrasts consistently in generated pairwise choices (Holm $p = 0.016$ each). Replacing AdamW with momentum SGD reproduces the ordered interaction in 3/3 tested seeds (Appendix~\ref{app:panels}).

The transplant and factorial results extend across five trait families. The 15 primary transplant cells select $w{+}m$ spanning Qwen (owl, cat, red, oak) and Llama (blue) (Appendix Figure~\ref{fig:extended_traits} and Table~\ref{tab:extended_traits}). Qwen blue appears in the observer-free and matched-factorial panels. Direct Qwen transplant responses range from red's negative seed mean through cat's near-zero mean to oak's positive mean (Table~\ref{tab:trait_behavior}), while the five-trait factorial measures route ordering and statistical resolution (Table~\ref{tab:factorial_5trait}).

\FloatBarrier

\textbf{Transport does not fix the behavioral outcome.} CNN block allocations vary tenfold across seeds. In a representative SmolLM2 cell, finite Shapley terms cancel by 90.9\% between source updates; its separate block-lineage panel identifies the two moments as the dominant pair. TinyLlama~\citep{zhang2024tinyllama} shows opposing finite contributions without a stable behavioral sign and does not rank transplant subsets. On Llama-3.2, direct source--control responses in the transplant panels remain weak and seed-heterogeneous, while the matched factorial resolves route-conditioned valuation at both eight and sixteen source-free updates. These cases show that ancestry can be transported even when behavioral value cancels, varies across seeds, or remains weak (Appendix Figure~\ref{fig:generalization}; per-system details in Appendix~\ref{app:panels}).

\section{Related Work}
\label{sec:related}

\textbf{Subliminal learning mechanisms.} Knowledge distillation \citep{hinton2015distilling} showed that a teacher's output distribution carries information beyond labels; subliminal learning \citep{cloud2025subliminal} extends this to behavioral dispositions invisible in the data. Subsequent work investigates how these subliminal signals enter the student's gradient: divergence tokens \citep{schrodi2026divergence}, steering-vector distillation \citep{blank2026steering}, LoRA amplification \citep{nief2026lora}, and stronger student encoding \citep{subliminalsteering2026}. We study what follows once the signal reaches the gradient port: its survival and expression within the trainer state.

\textbf{Adjoint methods and trajectory attribution.} Discrete-time adjoint analysis \citep{pontryagin1962} allows differentiating through momentum and optimizer buffers, as established in hyperparameter optimization \citep{maclaurin2015} and trajectory attribution. Traditional attribution methods trace sample influence via gradients or checkpoints \citep{koh2017understanding,pruthi2020tracin}; recent methods use approximate unrolling \citep{bae2024source}, fixed-state Adam-aware valuation \citep{ding2026inrun}, or explicit reverse-mode tracing through Adam/AdamW state \citep{anon2026adamw}. Building on this capability to differentiate through optimizer state, we isolate how future training continuation assigns behavioral value to a finite source-content contrast.

\textbf{Optimizer memory and order dependence.} Optimization literature recognizes that optimizer state acts as an implicit modification of the loss \citep{cattaneo2025memory} and steers later updates \citep{sweeney2026shuffle}. \citet{sevetlidis2026processtensor} demonstrated this memory effect directly by showing it collapses when optimizer buffers are reset. We use these memory dynamics to identify optimizer state as the causal carrier of source-specific ancestry and to separate physical transport from future behavioral valuation.

\section{Limitations and Conclusion}
\label{sec:limitations}

\textbf{Limitations.} The LLM panels span 135M--1.1B parameters and five Qwen and two Llama trait families. The transplant cells consistently select $w{+}m$, while behavioral magnitude and block-lineage allocation vary by system. Exact attribution requires the full trajectory and one adjoint solve per route; the compact midpoint predictor loses resolution when the source signal is small. The identity applies to gradient-port perturbations generally, with subliminal transfer distinguished by how the perturbation enters training.

\textbf{Conclusion.} A subliminal source can remain causally active after it leaves the training stream. Optimizer state carries that influence forward, while subsequent training determines whether it is expressed, cancelled, or reversed. The response identity, state surgery, and route predictions make transport and valuation separately measurable across language models, optimizers, traits, and non-LoRA vision systems. This separates what the trainer remembers from what the model eventually does: the same physical ancestry can persist while its behavioral value changes with the future route.

\begingroup
\small
\bibliographystyle{unsrtnat}
\bibliography{references}
\endgroup

\appendix

\section{Notation}
\label{app:notation}

\begin{table}[H]
\caption{Symbols used throughout the paper.}
\label{tab:notation}
\begin{center}
\small
\begin{tabular}{ll}
\toprule
\textbf{Symbol} & \textbf{Meaning} \\
\midrule
$S_t = (w_t, m_t, v_t, \tau_t)$ & complete trainer state at step $t$: parameters, first moment, \\
& second moment, auxiliary deterministic state (optimizer clock/schedule) \\
$x_t$ & training batch at step $t$ \\
$\alpha \in [-1,1]$ & source strength: $\alpha=-1$ neutral arm, $\alpha=+1$ trait arm, $\alpha=0$ midpoint \\
$P_\alpha$ & declared affine gradient-port path indexed by $\alpha$ \\
$G_t(S_t, x_t)$ & gradient computation (includes loss, backward pass, clipping input) \\
$U(S_t, g_t)$ & optimizer update map (SGD / momentum / AdamW step) \\
$O(S_T)$ & pre-declared behavioral observer applied at the horizon $T$ \\
$g_t$ & gradient handed from $G$ to $U$ (the \emph{gradient port}) \\
$d_t = \partial G_t / \partial \alpha |_{S}$ & source perturbation: gradient change caused by the teacher \\
& signal at step $t$, holding trainer state fixed \\
$q_t = \partial S_t / \partial \alpha$ & tangent (source ancestry): sensitivity of the full state to $\alpha$ \\
$A_t = S_t(+1)-S_t(-1)$ & finite ancestry chord between the two source arms \\
$p_t = (\partial O_T / \partial S_t)^\top$ & costate: sensitivity of the terminal observer to state $t$ \\
$\lambda_t = (\partial U / \partial g)^\top p_{t+1}$ & future value: what a gradient nudge at $t$ \\
& is worth to the final measurement \\
$\Wt = \langle \lambda_t, d_t \rangle$ & per-step work of the source against the future value \\
$R, R'$ & future training routes ($S^+$, $S^-$, $z_{\mathrm{null}}$) applied after a cut $E$ \\
$D_i = (y_i^- - y_i^+)/2$ & per-seed paired route contrast \\
$\SSEZERO$ & $\sum_i(\hat{y}_i - y_i)^2 / \sum_i y_i^2$; $\ll 1$ = signal captured, \\
& $\geq 1$ = no better than predicting zero \\
\bottomrule
\end{tabular}
\end{center}
\end{table}

\begin{table}[H]
\caption{Narrative terms used throughout the paper, in plain language.}
\label{tab:glossary}
\begin{center}
\small
\begin{tabular}{p{0.22\textwidth}p{0.68\textwidth}}
\toprule
\textbf{Term} & \textbf{Plain-language meaning} \\
\midrule
ancestry (source/physical/\allowbreak accumulated ancestry) & the trace the teacher signal leaves in the trainer state as it accumulates. The local tangent is $q_t=\partial_\alpha S_t$; a finite \emph{ancestry chord} is $A_t=S_t(+1)-S_t(-1)=\int_{-1}^{1}q_t(\alpha)\,d\alpha$ (\S\ref{sec:setup}) \\
relay (optimizer relay) & the two-step handoff by which ancestry is first held in optimizer state (first moment $m$ or momentum velocity) and only later written into parameters (\S\ref{sec:e3}) \\
transplant / reset / rescue & the three steps of the state-surgery protocol: \emph{transplant} copies a chosen subset of state blocks from a source-trained run into a control run; \emph{reset} sets every other block back to the control run's own values; \emph{rescue} is the shared source-free suffix of updates applied afterward, which is what converts a transplanted block's stored ancestry into a measurable effect (\S\ref{sec:e3}) \\
transport & the physical propagation of source ancestry through the trainer, independent of which behavioral readout is later applied to measure it (\S\ref{sec:setup}) \\
valuation & the sign and magnitude that a particular observer and a particular future training path assign to a given piece of ancestry; unlike transport, valuation depends on both (\S\ref{sec:identity}) \\
future value & the quantity $\lambda_t$: how much a gradient nudge at step $t$ would move the final behavioral measurement, computed backward from the observer (\S\ref{sec:setup}) \\
compatible entry & the initial stage at which a compatible teacher--student output interface and token mapping let a subliminal signal enter the student's gradients (\S\ref{app:e1e2}) \\
\bottomrule
\end{tabular}
\end{center}
\end{table}

\FloatBarrier
\section{Source Entry and Coordinate Access}
\label{app:e1e2}

\begin{figure}[H]
\centering
\includegraphics[width=\textwidth]{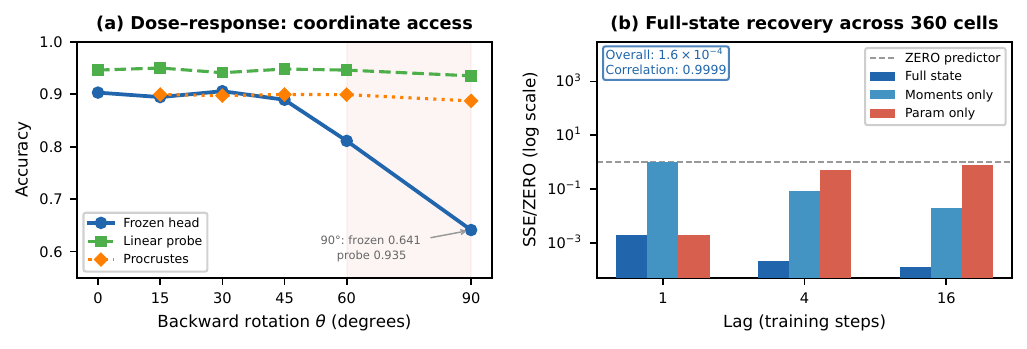}
\caption{\textbf{Source entry and coordinate access.} \captiona Backward rotation: frozen-head accuracy (solid) declines with $\theta$ while linear probe (dashed) stays flat---the source writes to different coordinates but the information persists. Procrustes (dotted) restores access. \captionb Full-state prediction: the joint predictor (dark blue) retains near-exact accuracy at all lags; parameter-only (red) degrades sharply by lag~16, and moments-only (light blue) by lag~1.}
\label{fig:birth}
\end{figure}

\textbf{Full-state prediction across 360 cells.} We apply a source pulse (a brief block of source-active updates) to the Qwen2.5-0.5B trainer (LoRA-r8, AdamW) and predict the complete state response---parameters, first moment, and second moment jointly---using the tangent recurrence (Eq.~\ref{eq:tangent}) with system-specific Jacobians, not a fitted model. Across 360 cells (step $\times$ state-block combinations) the joint prediction reaches $\SSEZERO = 1.65 \times 10^{-4}$ with correlation 0.9999. At lag~16 the full-state predictor maintains this accuracy, while a parameter-only predictor degrades to $\SSEZERO = 0.786$ and a moments-only predictor reaches 0.019 (Figure~\ref{fig:birth}b): near-exact tracking requires the full optimizer state.

\textbf{Backward-rotation dose response.} In a frozen-head MNIST experiment (classifying from the fixed output layer without retraining it), we rotate the hidden backward map by angle $\theta$ while preserving forward outputs, loss, gradient norm, and singular spectrum---changing \emph{where} the source writes without changing \emph{what} the network computes. Across $0^\circ$--$90^\circ$, frozen-head accuracy declines overall while a linear probe stays nearly flat, and unlabeled Procrustes alignment restores frozen-head accuracy---the information is present but written to different coordinates (Figure~\ref{fig:birth}a). A held-out $-60^\circ$ ancestry is predicted from three basis rotations by linear combination in the angular coordinate at $\SSEZERO < 10^{-6}$ (AdamW) and $< 10^{-12}$ (SGD), while wrong-source predictions score $\geq 1.0$.

\FloatBarrier

\section{Sustained Influence, Storage, and Observer}
\label{app:e6e7}

\begin{figure}[H]
\begin{center}
\includegraphics[width=\textwidth]{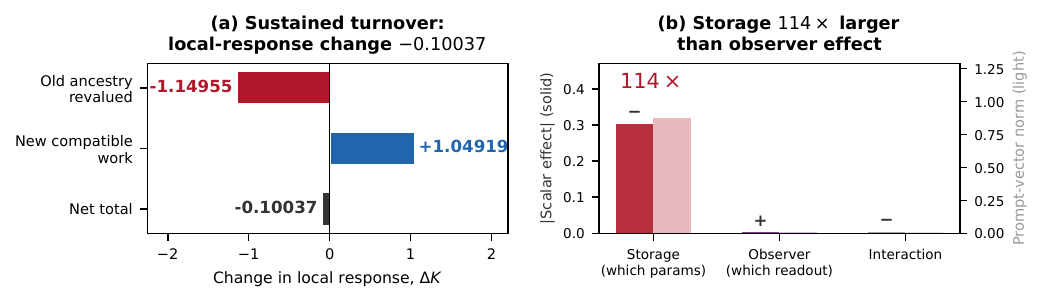}
\end{center}
\caption{\textbf{Sustained influence and storage--observer separation.} \captiona Old-ancestry revaluation ($-1.14955$) outweighs new source work ($+1.04919$), producing a net local-response change of $-0.10037$ at underlying precision. \captionb Storage is $114\times$ the observer effect. Solid bars show absolute scalar effects; light bars show prompt-vector norms.}
\label{fig:turnover}
\end{figure}

\textbf{Turnover decomposition.} In one Qwen2.5-0.5B/LoRA-r8/AdamW lineage, we extend the same lineage from $T=144$ to $T=160$ and evaluate the local $\alpha=0$ work. A net change of $-0.10037$ comprises $-1.14955$ from old ancestry being revalued and $+1.04919$ from new source work (Figure~\ref{fig:turnover}a; the displayed components are rounded). Sustained training is therefore a closed loop in which old and new contributions are continually re-weighed against each other, not a running total of isolated pulses.

\textbf{Storage--observer factorial.} In the same lineage, a $2 \times 2$ factorial crosses the stored states at $T=89$ and $T=96$ with $z_{\mathrm{null}}$ observers at the same two horizons, evaluated on eight held-out prompts. The storage effect ($-0.30291$) is $114\times$ the observer effect ($+0.00266$), with a small interaction ($-0.00345$; Figure~\ref{fig:turnover}b). This bounds these two specific interventions, not observers in general: because $\lambda_t$ depends on $O$, observers can disagree quantitatively or assign opposite signs to the same physical endpoint. The unrelated-observer rows in Table~\ref{tab:trait_behavior} realize the qualitative case on shared full-state endpoints, including within-seed sign changes across observers. Accordingly, the behavioral panels identify their observers, and we report physical transport separately from behavioral value.

\section{Proofs}
\label{app:proofs}

\subsection{Derivation of the Response Identity (Eq.~\ref{eq:cffl})}
\label{app:proof_cffl}

\textbf{Setup.} Fix the data order, randomness, optimizer code, and observer. Training is the deterministic recursion of Eq.~\ref{eq:trainer}:
\[
g_t = G_t(S_t, x_t; \alpha), \qquad S_{t+1} = U_t(S_t, g_t), \qquad O_T(\alpha) = O(S_T(\alpha)),
\]
where $G_t(S_t,x_t;\alpha)$ is the gradient selected by the declared affine gradient-port path $P_\alpha$. All dependence on the source strength $\alpha$ enters through this path. Assume each map is differentiable along the realized trajectory (the piecewise-smooth extension is Appendix~\ref{app:clipping}).

\textbf{Step 1: tangent recurrence.} Differentiate the composite $S_{t+1} = U_t(S_t, G_t(S_t,x_t;\alpha))$ with respect to $\alpha$. By the chain rule, with $q_t = \partial S_t / \partial \alpha$,
\[
q_{t+1}
= \frac{\partial U}{\partial S}\, q_t
+ \frac{\partial U}{\partial g} \frac{d g_t}{d \alpha}
= \frac{\partial U}{\partial S}\, q_t
+ \frac{\partial U}{\partial g}\!\left( \frac{\partial G}{\partial S}\, q_t + d_t \right),
\]
which is Eq.~\ref{eq:tangent}. The initialization does not depend on $\alpha$, so $q_0 = 0$.

\textbf{Step 2: general endpoint derivative.} Define the costate backward from the observer: $p_T = (\partial O / \partial S_T)^\top$ and, for $t < T$,
\[
p_t = \left(\frac{\partial U}{\partial S}\right)^{\!\top} p_{t+1} + \left(\frac{\partial G}{\partial S}\right)^{\!\top} \lambda_t,
\qquad
\lambda_t = \left(\frac{\partial U}{\partial g}\right)^{\!\top} p_{t+1},
\]
(Eqs.~\ref{eq:future_field} and~\ref{eq:costate}). By construction, $p_t^\top = \partial O_T / \partial S_t$ along the trajectory: it is the adjoint of the state-to-endpoint map. For a general protocol in which $\alpha$ could also enter the initialization, the observer, or the update map directly, the endpoint derivative is
\[
\partial_\alpha O_T
= p_0^\top\, \partial_\alpha S_0
+ \partial_\alpha O \big|_{S_T}
+ \sum_{t=0}^{T-1} p_{t+1}^\top\, \partial_\alpha U_t \big|_{S,g}
+ \sum_{t=0}^{T-1} \lambda_t^\top d_t.
\]
This follows by telescoping: writing $O_T$ as a composition of $T$ steps and collecting, for each step, the direct $\alpha$-dependence of that step weighted by the sensitivity of the endpoint to that step's output.

\textbf{Step 3: boundary conditions.} Our experimental protocol holds the initialization fixed ($\partial_\alpha S_0 = 0$), declares the observer independently of $\alpha$ ($\partial_\alpha O|_{S_T} = 0$), and keeps the optimizer code fixed ($\partial_\alpha U_t|_{S,g} = 0$); all source-coordinate dependence enters through $G$ at the gradient port. The first three terms vanish, leaving
\[
\partial_\alpha O_T = \sum_{t=0}^{T-1} \langle \lambda_t(\alpha), d_t(\alpha) \rangle.
\]

\textbf{Step 4: integration.} $O_T(\alpha)$ is absolutely continuous in $\alpha$ on $[-1,1]$ (a finite composition of smooth maps of $P_\alpha$), so by the fundamental theorem of calculus,
\[
O_T(+1) - O_T(-1) = \int_{-1}^{1} \partial_\alpha O_T \, d\alpha
= \int_{-1}^{1} \sum_{t=0}^{T-1} \langle \lambda_t(\alpha), d_t(\alpha) \rangle \, d\alpha,
\]
which is Eq.~\ref{eq:cffl}. \hfill $\square$

\textbf{Remark (port invariance).} The work $\langle \lambda_t, d_t \rangle$ is invariant to any state-independent invertible re-coordinatization of the gradient port: if $\tilde{g} = L g$, then $\tilde{d} = L d$, $\tilde{\lambda} = L^{-\top} \lambda$, and $\tilde{\lambda}^\top \tilde{d} = \lambda^\top d$. The decomposition is therefore a property of the implemented $G \to U$ interface, not of the units chosen for it.

\subsection{The Route Approximation (Eq.~\ref{eq:route_theorem})}

Let two routes $R, R'$ share the same pair of source-cut states, hence the same midpoint $S_E(0)$ and finite ancestry chord $A_E=S_E(+1)-S_E(-1)$. For $t \geq E$ the routes apply different training, producing different costates $p_E^R(0) \neq p_E^{R'}(0)$ at the cut. The endpoint response of route $R$ to the shared ancestry is, to first order in $\|A_E\|$,
\[
\Delta O_T^R \;=\; p_E^R(0)^\top A_E \;+\; o(\|A_E\|),
\]
because $p_E^\top = \partial O_T / \partial S_E$ is the linearization of the endpoint in the state at $E$. Subtracting,
\[
\Delta O_T^R - \Delta O_T^{R'} \;\approx\; \bigl(p_E^R(0) - p_E^{R'}(0)\bigr)^\top A_E.
\]
The shared factor $A_E$ cannot determine the sign of either term alone; the sign is carried by the route-dependent costates. The engineered factorial (\S\ref{sec:e4}) demonstrates continuation-dependent sign assignment with matched finite interventions, for which the exact identity provides the decomposition. The ordinary-route panel (\S\ref{sec:e4natural}) evaluates the first-order relation above directly at the sign and route-ordering level; magnitudes involve the neglected higher-order terms, and the finite-amplitude remainder is the curvature integral $\mathcal{R}_R = \int_{-1}^{1} [p_E^R(\alpha) - p_E^R(0)]^\top q_E(\alpha)\, d\alpha$ around the center state $\alpha = 0$.

\subsection{Remarks: Scope of the Identity}
\label{app:remarks}

Four observations describe the scope of the identity.

\textbf{Finite-horizon non-coalescence.} More generally, any matched continuation composed of injective complete-state updates preserves distinctions between cut states. For the standard non-AMSGrad AdamW update used here, fix the future batches, RNG stream, and schedule, and write
\begin{align*}
m_{t+1} &= \beta_1m_t+(1-\beta_1)g_t(w_t), \\
v_{t+1} &= \beta_2v_t+(1-\beta_2)g_t(w_t)^{\odot 2}, \\
w_{t+1} &= a_t w_t-\eta_tD_t(v_{t+1},\tau_t)m_{t+1},
\end{align*}
where $a_t=1-\eta_t\lambda_{\mathrm{wd}}$ and the known diagonal map $D_t$ includes bias correction and $\epsilon$ regularization. When $\beta_1$, $\beta_2$, and $a_t$ are nonzero, the preceding state is recovered explicitly:
\begin{align*}
w_t &= a_t^{-1}\!\left[w_{t+1}+\eta_tD_t(v_{t+1},\tau_t)m_{t+1}\right], \\
g_t &= G_t(w_t,x_t;\tau_t), \\
m_t &= \frac{m_{t+1}-(1-\beta_1)g_t}{\beta_1}, \qquad
v_t = \frac{v_{t+1}-(1-\beta_2)g_t^{\odot 2}}{\beta_2}.
\end{align*}
Thus each time-indexed update on $(w,m,v)$ is injective, and two distinct matched cut states in these optimizer blocks cannot coalesce after any finite continuation. The classical momentum-SGD update used here has the same property: from $u_{t+1}=\mu u_t+g_t(w_t)$ and $w_{t+1}=w_t-\eta_tu_{t+1}$, one recovers $w_t=w_{t+1}+\eta_tu_{t+1}$ and then $u_t=[u_{t+1}-g_t(w_t)]/\mu$ when $\mu\neq0$ \citep{maclaurin2015}. Plain SGD remains covered by the transport--valuation identity, but its parameter-only update need not be injective without additional conditions on the gradient map and step size. Non-coalescence is an exact-arithmetic statement about the complete optimizer state; it does not give a lower bound on ancestry magnitude or behavioral response.

\textbf{Experimental separation of the modules.} The interventions isolate the roles of the source term, optimizer map, and observer. The matched factorial varies source content under matched routes; the turnover decomposition separates old-ancestry revaluation from new source work ($-1.150$ vs.\ $+1.049$); state surgery distinguishes ancestry stored in parameter and optimizer blocks; and the storage--observer factorial varies the observer on shared physical endpoints.

\textbf{Observer relativity.} Two observers with endpoint rows $r$ and $-r$ induce identical observer-free trajectory descriptors but opposite signed responses whenever $r^\top q_T \neq 0$; no observer-free functional of the trajectory determines the sign. The storage--observer factorial and Table~\ref{tab:trait_behavior} provide the empirical counterpart: the same physical endpoint can receive quantitatively or qualitatively different values under different declared observers (\S\ref{app:e6e7}).

\textbf{Coordinate invariance and operational minimality.} Under a smooth invertible reparameterization $S' = \psi(S)$, the tangent and costate transform as $q' = D\psi\,q$ and $p' = D\psi^{-\top}p$, so $p'^\top q' = p^\top q$. Paired work and response totals are invariant, while the block names $w$, $m$, and $v$ depend on the chosen chart. If two reachable cut states produce the same declared physical outputs under every allowed source-free suffix, they are equivalent for that intervention family. The minimal causal state is the source-reachable set modulo this future indistinguishability. The $w{+}m$ result therefore identifies the sufficient block set in the implemented chart and source-free continuation family.

\section{The Full Computation}
\label{app:algorithm}

Algorithm~\ref{alg:cffl} computes the response derivative for one source strength $\alpha$. The forward sweep runs alongside ordinary training and maintains the tangent $q_t$ (Eq.~\ref{eq:tangent}); the backward sweep replays the trajectory in reverse and maintains the costate $p_t$ (Eq.~\ref{eq:costate}). Their meeting point is the per-step work $\Wt = \langle \lambda_t, d_t \rangle$, whose sum is $\partial_\alpha O_T$. The finite endpoint difference in Eq.~\ref{eq:cffl} is the integral of that returned sum over $\alpha$. This equality is exact; any finite-node quadrature used to evaluate it has a separate approximation error that must be assessed for the selected path. No empirical panel estimates the full finite-amplitude integral by fixed-node quadrature: the local-work panels evaluate $\alpha=0$, while finite-amplitude panels use matched interventions or finite Shapley decompositions. The compact midpoint predictor of Section~\ref{sec:e4natural} is evaluated explicitly as a first-order finite-amplitude approximation. Intervals containing persistent kinks or clipping events are integrated by matched finite interventions, avoiding possible distortions from pointwise derivatives (Appendix~\ref{app:clipping}).

\begin{algorithm}[H]
\caption{Full-state adjoint derivative at one source strength}
\label{alg:cffl}
\begin{algorithmic}[1]
\Require training data stream $\{x_t\}$, source strength $\alpha$, observer $O$, horizon $T$
\State \textbf{Forward sweep (with training):} $q_0 \gets 0$
\For{$t = 0, \dots, T-1$}
    \State $g_t \gets G_t(S_t, x_t; \alpha)$; \quad $S_{t+1} \gets U(S_t, g_t)$ \Comment{ordinary training step}
    \State $d_t \gets \partial G_t / \partial \alpha$ \Comment{source perturbation, state held fixed}
    \State $q_{t+1} \gets \frac{\partial U}{\partial S} q_t + \frac{\partial U}{\partial g}\big(\frac{\partial G}{\partial S} q_t + d_t\big)$ \Comment{JVPs; Eq.~\ref{eq:tangent}}
    \State record $(S_t, g_t, d_t)$
\EndFor
\State \textbf{Backward sweep:} $p_T \gets (\partial O / \partial S_T)^\top$
\For{$t = T-1, \dots, 0$}
    \State $\lambda_t \gets \big(\frac{\partial U}{\partial g}\big)^\top p_{t+1}$ \Comment{VJP; Eq.~\ref{eq:future_field}}
    \State $\Wt \gets \langle \lambda_t, d_t \rangle$ \Comment{per-step work}
    \State $p_t \gets \big(\frac{\partial U}{\partial S}\big)^\top p_{t+1} + \big(\frac{\partial G}{\partial S}\big)^\top \lambda_t$ \Comment{Eq.~\ref{eq:costate}}
\EndFor
\State \Return $\{\Wt(\alpha)\}_{t=0}^{T-1}$, \; $\partial_\alpha O_T = \sum_t \Wt(\alpha)$ \Comment{integrate over $\alpha$ for Eq.~\ref{eq:cffl}}
\end{algorithmic}
\end{algorithm}

\textbf{Cost and how it scales.} The method's scaling cost determines where it can be applied. Per source strength $\alpha$, the forward sweep adds a fixed number of JVP evaluations per update on top of ordinary training and the backward sweep adds a fixed number of VJP evaluations per update; the extra computation is therefore determined by those Jacobian products. The binding constraint is memory. The backward sweep replays the trajectory in reverse and therefore needs access to the recorded per-step quantities $(S_t, g_t, d_t)$, which is $O(T)$ storage in the horizon, each item the size of a full trainer state. At our scale (LoRA parameters on models up to 1.1B, horizons of order $10^2$ updates), exact replay was feasible. Gradient checkpointing trades recomputation for storage; segmented approximations of the kind SOURCE uses \citep{bae2024source} trade exactness for checkpoint-level storage. The $\alpha$-integral multiplies the cost by the number of evaluated source strengths; the required numerical resolution is a property of the selected path, not of the identity.

The interventional experiments set the larger compute cost in our runs: a block-transplant factorial forks matched states into eight whole-block combinations and re-runs the suffix for each seed and route. The costate predictor is cheaper than the full decomposition because it needs one center-state costate per route and no $\alpha$-sweep.

\section{Nonsmooth Events and Gradient Clipping}
\label{app:clipping}

The response identity is exact for smooth trainers and extends to the piecewise-smooth primitives used here through the conservative-Jacobian and path-differentiability calculus of \citet{bolte2021conservative}, building on classical piecewise-smooth automatic differentiation \citep{griewank2008}. For locally Lipschitz, path-differentiable primitives, the state path $\alpha \mapsto S_t(\alpha)$ is absolutely continuous, the tangent and costate recurrences hold with conservative-Jacobian selections from the same executed program, and the response integral holds almost everywhere in $\alpha$. The AdamW implementation used here is covered on reachable gradient histories, where $\hat{v}_t = 0$ implies $\hat{m}_t = 0$ and the $\epsilon$-regularized ratio is locally Lipschitz; the ambient AdamW map lacks local Lipschitz guarantees at $v=0, m \neq 0$. Continuous clipping crossings add no jump term because the state is continuous; a discontinuous branch would require explicit jump terms, and none occurs in our fixed-schedule trainers. Primal and dual sweeps use the same derivative selection at any kink. Appendix~\ref{app:adjoint_accuracy} checks their duality and finite-difference accuracy in active and inactive clipping regimes; crossings are evaluated one-sided rather than with a centered stencil. Intervals on which an event persists are integrated by matched finite interventions, avoiding possible distortions from pointwise derivatives across piecewise regions.

\section{Online Costate Approximation}
\label{app:online}

The approximations that would change the memory asymptotics of Appendix~\ref{app:algorithm} are truncation of the backward horizon and low-rank or sketched representations of the costate. We evaluate horizon truncation on the fully specified 42-route panel (\S\ref{sec:e4natural}): every tested shorter horizon remains below the full-horizon predictor in every seed, while the full horizon reproduces the reference predictions exactly (per-$H$ scores in Appendix~\ref{app:baselines}).

\section{Extended Relation to Prior Work}
\label{app:related}

\textbf{Divergence tokens and transferred value.} \citet{schrodi2026divergence} locate the carrier in the data: the token positions where teacher and reference distributions differ. Our decomposition addresses the complementary trainer-side question. Divergence identifies where the signal can enter; the costate describes how later training values the resulting gradient perturbation.

\textbf{Optimizer requirements.} \citet{blank2026steering} report that steering-vector distillation requires an adaptive optimizer in their LLM protocol. Our momentum-SGD replications reproduce the relay with velocity in place of Adam's first moment (\S\ref{sec:e3}, Appendix~\ref{app:panels}), showing that the transport path does not require adaptive scaling; optimizer choice changes its realized amplitude.

\textbf{Parameterization.} \citet{nief2026lora} find full fine-tuning eliminates transfer in their setting. Our MNIST experiments use no LoRA, and a descriptive matched-update-norm Qwen control includes LoRA, direct-adapter, and full-parameter arms (Table~\ref{tab:stepsize}). These results show that the mechanism does not require LoRA.

\textbf{Trainer-side channels.} \citet{madl2026channel} constrains where an auditable channel can sit. Our decomposition distinguishes trainer state from future-value sensitivity inside the training process. \citet{askin2026datamediated} compare data-mediated transfer conditions across task structure, prompt opportunity, and teacher/data channels. Their emphasis on dataset structure is complementary to our optimizer-centric account and may help explain why physical transport need not yield stable behavior.

\textbf{Larger-model and agent settings.} \citet{anon2026ratios} quantify subliminal behavioral transfer in 7B language-model distillation, and \citet{anon2026unsafe} study unsafe-behavior transfer in AI-agent distillation. These results broaden the settings in which the phenomenon has been measured; our experiments address the complementary trainer-state mechanism under controlled interventions.

\textbf{Seed-induced uniqueness and representational substrate.} \citet{okatan2025seed} report that transfer strength tracks alignment inside a trait-discriminative subspace rather than global representational similarity, so that students differing only in initialization seed leak substantially less than same-seed students even at global CKA above $0.9$. That result concerns the representational substrate a trait needs in order to be readable; ours concerns where the trait is physically held while training continues. The two are complementary: seed-specific geometry may help explain why the allocation of source response across parameters and optimizer slots varies widely across seeds (Appendix~\ref{app:seeds}).

\textbf{Metric circularity.} \citet{chauhan2026covert} identify a shared-denominator artifact in regressions involving log probability ratios. Our primary estimand is a matched source--route interaction, and the generated-choice panel evaluates its route ordering without that regression (\S\ref{sec:behavior_validation}).

\textbf{Compatible output heads.} \citet{brockers2026noise} show that shared output-layer structure between teacher and student is a precondition for transfer. Our backward-rotation experiment (\S\ref{app:e1e2}) builds on this: rotating the hidden backward map by angle $\theta$ while preserving forward outputs shows where the source writes rather than what it writes, and a Procrustes realignment restores access.

\textbf{Temporal localization and liminal training.} \citet{yanagisawa2025liminal} localize trait acquisition to an early nonlinear transition and mitigate it with an annealed KL regularizer. Their mitigation acts during source exposure. Our source-position ablation asks whether the transport--valuation topology recurs when the source appears later: all four tested positions preserve the route ordering at 77--100\% of the reference amplitude (Figure~\ref{fig:ancestry}c).

\textbf{Emergent misalignment.} \citet{betley2025emergent} show that fine-tuning on a narrow task---writing insecure code---can induce broadly misaligned behavior across unrelated domains. There the disposition is not an explicit target but the misaligning source data is overtly present. That work studies why a narrow source generalizes broadly; our decomposition studies how update contributions are stored and later valued.

\textbf{Non-semantic transfer and task arithmetic.} \citet{chiang2022transferability} showed that pre-training transfer can arise from artificial, non-semantic properties of the data---the broader question of what training data can transmit besides its overt content. Trait vectors obtained by weight-space merging \citep{anon2025personality}---following the task-arithmetic line, in which differences of fine-tuned weights are added and negated as vectors \citep{ilharco2023task}---place behavioral traits in parameter space; our transplant results show that ancestry can also reside transiently in optimizer slots before it becomes parameter-visible.

\textbf{Optimizer memory.} \citet{sevetlidis2026processtensor} measure non-Markovian memory in SGD training directly---via a back-flow-of-distinguishability witness---and find that it scales with momentum and collapses under a causal break that resets optimizer state, which agrees with our transplant results on where the memory physically sits. \citet{cattaneo2025memory} give a complementary theoretical account, showing that an optimizer's accumulated history acts as an implicit modification of the loss being descended, so later updates are steered by stored state and not by the current gradient alone.

\textbf{Order dependence.} \citet{sweeney2026geometry} shows that which of two training phases comes first changes the final transfer outcome, and that the change is predicted by the Lie-bracket commutator of the two update fields. Their object is the geometric non-commutativity of update order; ours is the causal valuation of a fixed perturbation by its continuation. Both say that when an update happens relative to the rest of training is part of what that update does.

\textbf{Unlearning persistence.} \citet{hu2025jogging} show that apparently unlearned content can return after benign follow-up fine-tuning, and \citet{xu2026reversibility} find that forgetting is often suppression near the output rather than erasure. \citet{georgiev2025attribute} use data attribution to predict the retrained-without-$x$ model and then fine-tune toward it. Our first-moment release result is a trainer-state analogue: ancestry can have zero forward-visible effect at the cut and still be released by later source-free updates.

\textbf{Persistence across subsequent training.} Backdoored behaviors can survive supervised fine-tuning, RL, and adversarial training \citep{hubinger2024sleeper}, and \citet{cui2025persistent} engineer poisoned gradients to persist through continual fine-tuning. Those studies examine intentionally persistent behavior; our experiments characterize how source-induced ancestry is relayed by trainer state.

\textbf{Other related methods.} Feedback alignment showed learning survives replaced backward maps \citep{nokland2016direct,refinetti2021align,garg2022random}; critical learning periods \citep{achille2019critical} established that transient early-training deficits become permanently harder to correct. LoRA and its variants \citep{hu2022lora,zhu2024asymmetry,hao2024flora,hayou2024loraplus} provide the parameterization used in our Qwen experiments.

\section{Validation Panels}
\label{app:panels}

\textbf{Generated-choice panel.} Seven seeds run the factorial with a pairwise-choice prompt grid held out from all geometry construction, scored one-shot at temperature 0. Both adjacent route contrasts are positive in generated choices in 7/7 seeds (one-sided sign test, Holm-adjusted $p = 0.016$ each). The panel validates the ordering the likelihood observer induces over routes, not the calibration of its magnitudes.

\textbf{Resolution of the $n = 7$ sign test.} A unanimous 7/7 result has one-sided raw $p = 0.0078$; Holm adjustment over the two adjacent contrasts gives $p = 0.016$.

\textbf{Momentum-SGD extension of the factorial.} The same Qwen protocol with AdamW replaced by momentum SGD (momentum 0.9, learning rate chosen from surface-task progress) reproduces the ordered factorial interaction in 3/3 seeds, with the same ordering recovered on an independent holdout bank. Each adjacent contrast has one-sided raw $p = 0.125$; Holm adjustment over the two contrasts gives $p = 0.25$ each, so this panel is descriptive. The mechanism can therefore run with momentum velocity in place of Adam moments.

\textbf{Two-trait matched factorial.} A matched Qwen factorial uses independently written 128-row native sources for the owl and blue traits under shared prompts, geometry, and horizons. The blue source reaches 6/7 seeds on both adjacent route contrasts, with a single seed reversing both; that reversal does not recur under the matched owl source in the same cell. The separate source-scale titration in Appendix~\ref{app:baselines} is consistent with comparable small-corpus scales falling in the low-signal regime.

\textbf{Optimizer-slot lineage.} The panel uses one seed per model, with two nonterminal cuts and three routes per cut. Each source-induced ancestry is split according to where it lives in the AdamW state---the parameters $w$, the first moment $m$, or the second moment $v$---and each block is propagated through the same future separately. We score a block $c$ by its \emph{block-restricted recovery error} $\SSEZERO(c\text{-only})$, normalized by the error of the zero predictor. These nonnegative prediction errors are not additive mass fractions: interactions between blocks are not credited to any single block. Separate signed allocations can be negative. In Qwen the dominant pair is the parameters together with the first moment, reaching $\SSEZERO \approx 5 \times 10^{-4}$, whereas in SmolLM2-135M it is the two moments instead. The dominant pair therefore differs between these two cells.

\textbf{SmolLM2 cancellation cell.} A finite Shapley decomposition over the three source-active updates assigns $-1.21 \times 10^{-3}$ to the first update and $+6.0 \times 10^{-4}$ and $+4.1 \times 10^{-4}$ to the second and third, against a net endpoint response of $-2.0 \times 10^{-4}$. In other words, 90.9\% of the gross allocation cancels between updates, so the near-zero net response results from the cancellation of substantial opposing contributions rather than the absence of transport. The allocation is a signed decomposition along the declared path, not a fraction of physical mass, which is why individual terms can be negative.

\textbf{Boundary-case response summaries.} In TinyLlama, finite interventions attribute 62.7\% to 83.6\% of the gross finite allocation to cancellation, depending on the cell, which is why the net behavioral response has no stable sign across seeds. In Llama-3.2-1B the seed-mean $\pm$ sample-SD animal-preference response is $-0.012 \pm 0.013$ under AdamW and $-0.005 \pm 0.005$ under momentum SGD across three seeds each, and the matched color-preference response is $+0.016 \pm 0.025$: weak and seed-heterogeneous in every case, even though every physical stage of the mechanism is observed in the same runs.

\section{Route-Aware Baselines and Random-Plane Control}
\label{app:baselines}

For seed $s$, $\operatorname{MSE}_s$ is computed over the concatenated prompt-level response vectors from its six routes. The baseline-margin score is $G_s^\ast = \min_b\{\operatorname{MSE}_s(b)-\operatorname{MSE}_s(\widehat{\Delta})\}$ over the ZERO, source-cut linear, and route-constant comparators. Thus $G_s^\ast>0$ exactly when the costate predictor beats all three.

\textbf{Receding-horizon predictors.} Each variant propagates the same finite ancestry chord $A_E$ through only the first $H$ of the eight suffix updates and reads the observer at the post-$H$ center state, which is the forward dual of an $H$-step online costate (Appendix~\ref{app:online}). Mean $G^\ast_s$ ($\times 10^{-3}$) is $-2.13$ at $H = 1$, $-1.55$ at $H = 2$, $-0.37$ at $H = 4$, and $+0.15$ at $H = 8$; the count of seeds with $G^\ast_s > 0$ is 0/7, 0/7, 0/7, and 7/7. Truncation degrades the predictor smoothly, but no truncated horizon beats the three primary baselines.

\textbf{Forward-only alignment.} The alignment predictor scores each route by $\sum_t \langle A_E^{(w)}, \bar g_t \rangle$, where $A_E^{(w)}$ is the parameter block of the cut-state ancestry and $\bar g_t$ is that route's center-state parameter gradient. This route-aware quantity requires no backward pass. It matches observed route-mean signs in 17 of 42 cells (per-seed range 0/6 to 6/6), and its per-seed Spearman correlations against observed endpoints are inconsistent in sign. Route-dependent forward signal alone does not recover the valuation.

\textbf{Schedule statistics.} Leave-one-route-out sibling means, nearest-neighbor, and kernel-weighted predictors are constructed from the observed endpoints of the other five routes in the same seed. The costate predictor has lower error in 6/7 seeds for the sibling-mean and kernel predictors and in 7/7 for nearest-neighbor. Exact two-sided Mantel permutation tests give $p \ge 0.026$ with association signs mixed across seeds; they do not show a consistent schedule--endpoint association.

\textbf{Random-plane control.} The control in \S\ref{sec:e4} draws one fixed random direction, orthogonalizes it against both $\hat\phi$ and the corpus gradient contrast $u$, and rescales it to $\|\phi\|$. Across seeds, the adjacent contrasts are $|d_1| = 0.136 \pm 0.071$ and $|d_2| = 0.149 \pm 0.056$ (mean $\pm$ sample SD; Cohen $|d_z| = 1.9$ and $2.7$), both sign-coherent in 7/7 seeds. The realized sign is opposite to the observer-informed plane. Under a random plane the assignment of the $S^\pm$ labels to the two half-spaces carries no observer meaning, so only sign \emph{coherence} across seeds is interpretable.

\textbf{Source-scale titration.} The ladder uses fixed paired-row subsamples of 128, 256, 512, and 1024 examples from the full 1831-row owl corpus; route generation, geometry, optimizer, horizons, and observer remain fixed. Mean $G^\ast_s$ ($\times 10^{-3}$) is $-0.69$ (128), $+0.07$ (256), $-0.03$ (512), $+0.15$ (1024), and $+0.08$ (1831). The independently written native 128-row protocol gives $-1.71$. Between 256 and 512 rows the predictor is near the zero-margin boundary and seed-unstable, where the route-dependent signal approaches the baseline/noise floor.

\section{Implementation Checks}
\label{app:validity}

Appendix~\ref{app:adjoint_accuracy} reports numerical derivative checks for AdamW and momentum SGD.

\textbf{Matched state in the transplant.} The surgery holds auxiliary state $\tau$, data order, and the RNG stream fixed across arms, so it isolates changes in $(w,m,v)$ under a common continuation.

\section{Extended Figures and Tables}
\label{app:seeds}

\begin{figure}[H]
\begin{center}
\includegraphics[width=\textwidth]{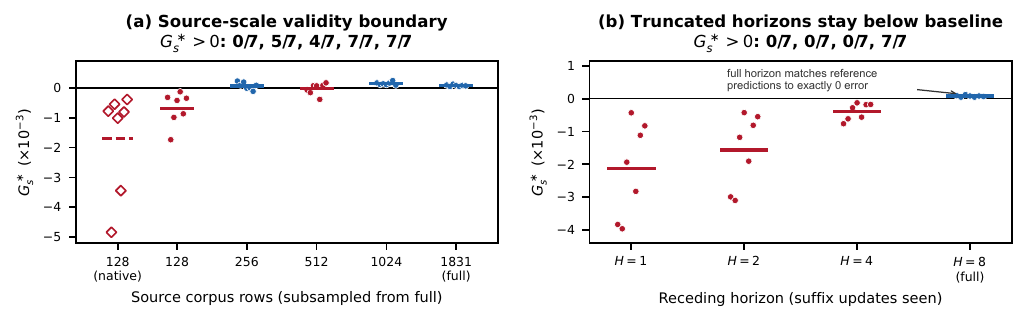}
\end{center}
\caption{\textbf{Validity domain of the compact chord predictor} (\S\ref{sec:e4natural}). Each dot is one seed's baseline-margin score $G^\ast_s$ (positive = the costate predictor beats ZERO, source-only, and route-constant competitors); horizontal bars are seed means. \captiona Source-scale titration: fixed paired-row subsamples from the full 1831-row corpus, with all other protocol elements unchanged. The predictor is below the baselines at 128 rows, transitions through a low-signal band at 256--512, and wins in 7/7 seeds at 1024 and above. Open diamonds: the independently generated native 128-row protocol is also below the baselines. \captionb Receding-horizon ladder on the same ordinary-route panel: truncating the propagated future to $H \in \{1,2,4\}$ of the 8 suffix updates gives negative margin in every seed.}
\label{fig:validity}
\end{figure}

\begin{figure}[H]
\centering
\includegraphics[width=\textwidth]{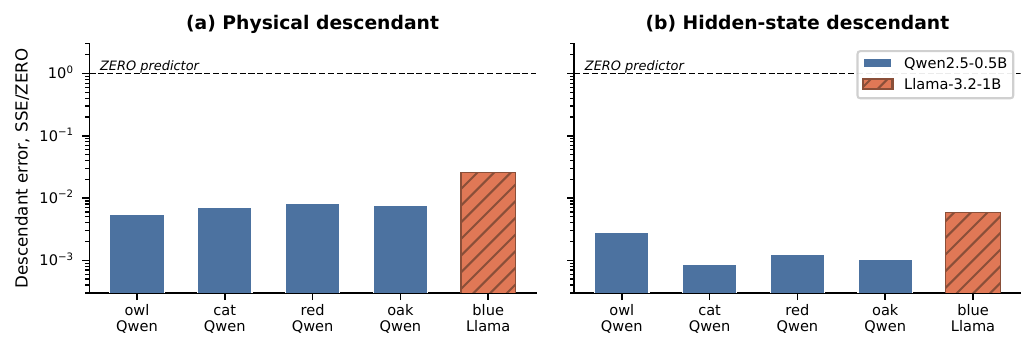}
\caption{\textbf{Five-trait transplant across Qwen2.5-0.5B and Llama-3.2-1B.} The $w{+}m$ transplant reproduces the full descendant in all 15 primary seed--trait cells: Qwen owl, cat, red, and oak, plus Llama blue. \captiona Physical descendant error. \captionb Hidden-state descendant error.}
\label{fig:extended_traits}
\end{figure}

\begin{figure}[H]
\centering
\includegraphics[width=\textwidth]{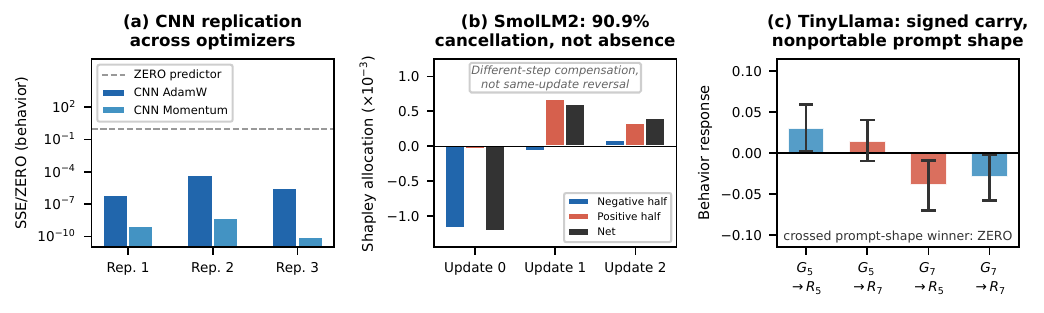}
\caption{\textbf{Generalization and boundary.} \captiona CNN replication across optimizers; block allocations are seed-dependent. \captionb One SmolLM2 cell: 90.9\% step-wise cancellation, not absence. \captionc In TinyLlama, response signs follow the transplanted gradient history, but the prompt-level response pattern does not transfer. Intervals are paired-prompt 95\% bootstrap intervals.}
\label{fig:generalization}
\end{figure}

\FloatBarrier

\begin{table}[H]
\caption{Separate Qwen common-ancestry route-replacement cohort ($n=7$): paired contrasts $D_i = (y_i^- - y_i^+)/2$, where $y_i^\pm$ is seed $i$'s response under $S^\pm$. All positive; two-sided sign-test $p = 0.016$.}
\begin{center}
\small
\begin{tabular}{lccccccc}
\toprule
\textbf{Replicate} & 1 & 2 & 3 & 4 & 5 & 6 & 7 \\
\midrule
$D_i$ & 0.243 & 0.223 & 0.528 & 0.303 & 0.465 & 0.373 & 0.397 \\
\bottomrule
\end{tabular}
\end{center}
\end{table}

\begin{table}[H]
\caption{Source-position ablation: route topology recurs across positions.}
\begin{center}
\small
\begin{tabular}{lcccc}
\toprule
\textbf{Position} & $D$ & \textbf{\% of ref.} & \textbf{Prompt cos.} & \textbf{64/64 pos.} \\
\midrule
40--47 & 0.320 & 77\% & 0.999 & \checkmark \\
80--87 (ref.) & 0.415 & 100\% & --- & \checkmark \\
120--127 & 0.367 & 89\% & 0.998 & \checkmark \\
200--207 & 0.342 & 82\% & 0.999 & \checkmark \\
\bottomrule
\end{tabular}
\end{center}
\end{table}

\begin{table}[H]
\caption{SmolLM2 source-step Shapley allocations. Cross-step cancellation: 90.9\%.}
\begin{center}
\small
\begin{tabular}{lccc}
\toprule
& \textbf{Update 0} & \textbf{Update 1} & \textbf{Update 2} \\
\midrule
Negative half & $-1.17 \times 10^{-3}$ & $-7.1 \times 10^{-5}$ & $+8.5 \times 10^{-5}$ \\
Positive half & $-4.6 \times 10^{-5}$ & $+6.7 \times 10^{-4}$ & $+3.3 \times 10^{-4}$ \\
Net & $-1.21 \times 10^{-3}$ & $+6.0 \times 10^{-4}$ & $+4.1 \times 10^{-4}$ \\
\bottomrule
\end{tabular}
\end{center}
\end{table}

\begin{table}[H]
\caption{TinyLlama finite anatomy: stable early/late opposition with 62--84\% cancellation.}
\begin{center}
\small
\begin{tabular}{lccc}
\toprule
\textbf{Replicate} & \textbf{Net chord} & \textbf{Cancel.} & \textbf{Polarity} \\
\midrule
1 & $+0.028$ & 63\% & Early$-$/Late$+$ \\
2 & $+0.020$ & 84\% & Early$-$/Late$+$ \\
3 & $+0.007$ & 62\% & Early$+$/Late$-$ \\
\bottomrule
\end{tabular}
\end{center}
\end{table}

\begin{table}[H]
\caption{CNN AdamW: block allocations vary widely across seeds. Param and $m_1$ are signed response shares normalized by the net response; omitted block and interaction terms are also signed, so the displayed shares need not sum to one.}
\begin{center}
\small
\begin{tabular}{lccc}
\toprule
\textbf{Replicate} & \textbf{$\SSEZERO$} & \textbf{Param share} & \textbf{$m_1$ share} \\
\midrule
1 & $7.8 \times 10^{-7}$ & 0.973 & 0.114 \\
2 & $5.4 \times 10^{-5}$ & 0.073 & 0.744 \\
3 & $3.0 \times 10^{-6}$ & 0.139 & 0.891 \\
\bottomrule
\end{tabular}
\end{center}
\end{table}

\begin{table}[H]
\caption{CNN momentum SGD: relay topology is stable across seeds. Src-cut is the source-cut response norm in the experiment's native units. Vel. and Param. are signed fractions of the terminal response and sum to one.}
\begin{center}
\small
\begin{tabular}{lcccc}
\toprule
\textbf{Replicate} & \textbf{$\SSEZERO$} & \textbf{Src-cut} & \textbf{Vel.} & \textbf{Param.} \\
\midrule
1 & $8.8 \times 10^{-10}$ & 4.22 & $+1.045$ & $-0.045$ \\
2 & $4.9 \times 10^{-9}$ & 2.02 & $+1.125$ & $-0.125$ \\
3 & $9.0 \times 10^{-11}$ & 8.50 & $+1.065$ & $-0.065$ \\
\bottomrule
\end{tabular}
\end{center}
\end{table}

\FloatBarrier

\section{Experimental Protocols}
\label{app:protocols}

\subsection{Qwen LLM Protocol}

Qwen2.5-0.5B-Instruct is trained with rank-8 LoRA and AdamW (learning rate $2 \times 10^{-4}$, $\beta=(0.9,0.999)$). The paired owl/neutral data contain approximately 1,831 bare-number completions per arm, generated by an owl LoRA teacher and the base-neutral teacher; training uses bare text without a chat template. In the factorial, a shared prefix ends at update 79, source training occupies updates 80--87, and source-free routes occupy updates 88--95, with terminal $T=96$. ``Neutral'' denotes the matched source-coordinate control, not absence of training. Behavior is the target-minus-reference conditional log-likelihood on a frozen prompt bank. The transplant protocol uses 40 prefix, 4 source, and 4 source-free updates. The momentum extension uses coefficient 0.9.

\subsection{Canonical Source Path}
\label{app:source_path}

For a declared route $R$, let $g_{O,t}^R(S)$ and $g_{N,t}^R(S)$ be its owl and neutral corpus gradients at state $S$. In the common-cut factorial of \S\ref{sec:e4}, these source-window gradients are shared across $R$ and only the suffix map changes after cut $E$. The source coordinate enters at the optimizer's gradient port as
\[
g_t^{\alpha,R}(S)
= \frac{g_{O,t}^R(S)+g_{N,t}^R(S)}{2}
+ \alpha\,\frac{g_{O,t}^R(S)-g_{N,t}^R(S)}{2},
\qquad -1\leq\alpha\leq1.
\]
Thus $\alpha = +1$ recovers the owl arm and $\alpha = -1$ the neutral arm within the same route, consistent with the integral bounds of Eq.~\ref{eq:cffl}. The matched finite contrast defines a source-content differential within a compatible interface, independent of literal source absence. This is the declared path for all $\alpha$-integrals; alternative paths sharing the same endpoints leave totals unchanged but redistribute per-step attribution (Appendix~\ref{app:routes}).

\subsection{Route Construction}
\label{app:routes}

\textbf{Engineered diagnostic routes.} $S^\pm$ span $e_\pm \propto \hat\phi \pm \hat u$ (with a shared completion direction $z$), where $\phi$ is the frozen behavior-readout direction from the declared prompt bank and $u$ the owl--neutral corpus gradient contrast; $z_{\mathrm{null}}$ is orthogonal to both. The geometry derives strictly from readout and gradient contrast directions, without referencing the costate or endpoint outcome. The random-plane control (\S\ref{sec:e4}) replaces $\hat\phi$ with a fixed random direction orthogonalized against $\hat\phi$ and $u$ and rescaled to $\|\phi\|$.

\textbf{Ordinary routes.} Each seed uses six random future data schedules, and all 42 routes are retained. The source-cut linear comparator reads the ancestry chord at $E$ without suffix propagation; the route-constant comparator assigns the across-route mean costate prediction to every route.

\textbf{Attribution-path robustness.} Two exact nested coalition traversals of the eight source updates (ascending, descending) share identical endpoints: the route topology $S^+ < z_{\mathrm{null}} < S^-$ is path-independent, while update-level work allocation shifts between paths (sign agreement 0.875--1.0). Endpoint totals are path-invariant; temporal allocation is reported relative to the declared path.

\textbf{Source-window ablation.} The three additional source blocks occupy updates 40--47, 120--127, and 200--207 (reference: 80--87). Each replicates the signed route topology, retains 77--89\% of the reference amplitude, and matches the reference prompt-vector direction at cosine 0.998--0.999. The consistently positive prompt contrasts evaluate repeated probes across the four positions, serving as descriptive within-seed measurements.

\textbf{Schedule-statistics baselines and Mantel tests.} For each held-out route, three predictors are built from the other five routes of the same seed: their observed-endpoint mean, the endpoint of the schedule-nearest sibling (mean absolute difference of per-example scheduled times), and a kernel-weighted sibling mean. These baselines use observed sibling endpoints, whereas the costate predictor does not. The costate predictor has lower error in 6/7 seeds for the mean and kernel predictors and in 7/7 for nearest-neighbor on the full-corpus panel. Exact Mantel permutation tests find no consistent association between schedule distance and endpoint distance (two-sided $p \ge 0.026$, signs mixed across seeds).

\subsection{MNIST Protocol}

A fixed compatible auxiliary head supplies task-unrelated training targets, while the inherited class head remains frozen for behavior readout. The MLP uses no LoRA; the CNN and optimizer extensions use three independent seeds.

\subsection{Observer-Free Prediction Panel: Operational Specification}
\label{app:obsfree}
Each system uses three independent seeds and four ordinary suffix routes over a matched future-minibatch multiset; routes are repeated measures within a seed, and seeds are the replication unit. Horizons in updates (prefix/source/source-free future) are $40/4/4$ for Qwen, $8/4/4$ for SmolLM2, and $12/4/4$ for Llama-3.2. All three use AdamW (learning rate $2 \times 10^{-4}$, $\beta=(0.9,0.999)$) and rank-8 LoRA.

\textbf{Comparison spaces and normalization.} Three inner-product spaces are scored. \emph{Full state} sums Euclidean inner products over the implemented LoRA parameter, first-moment, and second-moment tensors; prediction and target share the same adapter parameterization. \emph{Merged weights} uses the exact low-rank tangent $\delta(BA)=B\,\delta A+\delta B\,A$ of each adapted projection. \emph{Hidden} concatenates final-token states from the embedding and every transformer block over a fixed neutral probe bank; its target is the central finite difference under the declared source perturbation. In each space, ZERO is the summed target energy over the four routes, so $\SSEZERO$ is comparable across spaces of different raw scale; the route-centered variant subtracts the across-route mean from prediction and target before scoring.

\textbf{Comparators.} The source-only comparator repeats the source-cut chord for every route; the route-constant comparator repeats the across-route mean tangent prediction.

\textbf{Per-seed raw errors.} Table~\ref{tab:obsfree_raw} reports the un-normalized tangent-prediction error (SSE, in each space's own inner-product units), its ZERO denominator, and the ratio, per seed. The figure reports seed means of the ratios.

\begin{table}[H]
\centering
\caption{Observer-free panel: per-seed raw tangent-prediction error (SSE), ZERO denominator (summed target energy over the four routes), and their ratio, in each comparison space.}
\label{tab:obsfree_raw}
\scriptsize
\setlength{\tabcolsep}{3.5pt}
\begin{tabular}{ll ccc ccc ccc}
\toprule
& & \multicolumn{3}{c}{full state} & \multicolumn{3}{c}{merged weights} & \multicolumn{3}{c}{hidden} \\
\cmidrule(lr){3-5}\cmidrule(lr){6-8}\cmidrule(lr){9-11}
System & Rep. & SSE & ZERO & ratio & SSE & ZERO & ratio & SSE & ZERO & ratio \\
\midrule
Qwen2.5-0.5B & 1 & $2.0{\times}10^{-7}$ & 0.42 & $4.7{\times}10^{-7}$ & $7.6{\times}10^{-8}$ & 0.26 & $3.0{\times}10^{-7}$ & 0.29 & $4.7{\times}10^{3}$ & $6.2{\times}10^{-5}$ \\
 & 2 & $3.6{\times}10^{-7}$ & 0.48 & $7.6{\times}10^{-7}$ & $1.2{\times}10^{-7}$ & 0.29 & $4.2{\times}10^{-7}$ & 0.18 & $4.2{\times}10^{3}$ & $4.3{\times}10^{-5}$ \\
 & 3 & $2.7{\times}10^{-7}$ & 0.49 & $5.5{\times}10^{-7}$ & $1.0{\times}10^{-7}$ & 0.30 & $3.4{\times}10^{-7}$ & 0.29 & $3.8{\times}10^{3}$ & $7.7{\times}10^{-5}$ \\
\midrule
SmolLM2-135M & 1 & $3.6{\times}10^{-6}$ & 0.31 & $1.2{\times}10^{-5}$ & $1.8{\times}10^{-6}$ & 0.25 & $7.2{\times}10^{-6}$ & 0.32 & $8.9{\times}10^{3}$ & $3.6{\times}10^{-5}$ \\
 & 2 & $2.0{\times}10^{-6}$ & 0.26 & $7.9{\times}10^{-6}$ & $1.0{\times}10^{-6}$ & 0.22 & $4.6{\times}10^{-6}$ & 0.32 & $9.0{\times}10^{3}$ & $3.6{\times}10^{-5}$ \\
 & 3 & $1.8{\times}10^{-6}$ & 0.24 & $7.7{\times}10^{-6}$ & $1.1{\times}10^{-6}$ & 0.21 & $5.3{\times}10^{-6}$ & 0.31 & $1.4{\times}10^{4}$ & $2.2{\times}10^{-5}$ \\
\midrule
Llama-3.2-1B & 1 & $8.0{\times}10^{-5}$ & 0.71 & $1.1{\times}10^{-4}$ & $3.3{\times}10^{-5}$ & 0.57 & $5.8{\times}10^{-5}$ & 9.84 & $1.9{\times}10^{3}$ & $5.1{\times}10^{-3}$ \\
 & 2 & $5.1{\times}10^{-5}$ & 0.74 & $6.9{\times}10^{-5}$ & $1.0{\times}10^{-5}$ & 0.60 & $1.7{\times}10^{-5}$ & 0.85 & $1.3{\times}10^{3}$ & $6.6{\times}10^{-4}$ \\
 & 3 & $1.1{\times}10^{-4}$ & 0.82 & $1.4{\times}10^{-4}$ & $2.4{\times}10^{-5}$ & 0.69 & $3.5{\times}10^{-5}$ & 6.07 & $2.4{\times}10^{3}$ & $2.5{\times}10^{-3}$ \\
\bottomrule
\end{tabular}
\end{table}

\subsection{Llama-3.2 Protocol}
\label{app:llama}

Llama-3.2-1B-Instruct~\citep{meta2024llama32,unsloth2023} uses model-native paired owl/neutral and blue/neutral sources. The owl and neutral teachers generate 128 paired rows; on a disjoint qualification bank their contrast is $+21.2$ on the animal observer versus $+0.41$ on an unrelated color observer, and a lexical scan finds no target or control animal/color word. The transport and transplant panels use 12 prefix, 4 source, and 4 source-free updates. Three seeds are used for each panel: observer-free prediction uses four ordinary routes per seed; the AdamW transplant tests all eight $w/m/v$ hybrids under natural and reversed suffixes; and the momentum-SGD transplant uses $\mu=0.9$ with state $(w,m_{\mathrm{vel}})$. Its learning rate is not progress-matched to AdamW, so this panel tests the relay topology across distinct optimizers. In the behavioral panels, a response is called resolved only when its prompt-level 95\% bootstrap interval excludes zero and its magnitude exceeds replay, unrelated-observer, and numerical envelopes; a predicted sign must also be stable under numerical refinement.

\textbf{Transplant results.} Under AdamW with the owl source, $w{+}m$ beats every proper subset across all tested seeds ($\SSEZERO = 0.030$ physical, 0.011 hidden; $v$-only $\approx$ ZERO), and the $m$-only transplant is exactly forward-invisible at the cut, reaching merged-weight norm 0.061 after the first source-free update and 0.196 at the terminal state, with the hidden norm increasing from 3.7 to 9.8 over the same interval. Under momentum-SGD, velocity-only ancestry is again invisible at the cut and is released by subsequent updates in every seed.

\textbf{Behavioral factorial.} The matched $80/8/8$ source$\times$route design is repeated over 12 independent runs. The main likelihood observer gives $\Delta_{S^+} = -0.131 \pm 0.050$, $\Delta_{z_{\mathrm{null}}} = -0.017 \pm 0.035$, and $\Delta_{S^-} = +0.090 \pm 0.031$ (mean $\pm$ sample SD), with both adjacent contrasts positive across all runs (one-sided sign test, Holm-adjusted $p = 0.000488$ each). A held-out likelihood observer preserves both contrast signs in 12/12 runs. Mean physical descendant norms are 1.513, 1.507, and 1.513 across the three routes, and the source-cut behavioral response is $-0.001 \pm 0.026$.

\textbf{Longer-horizon factorial.} A separate paired panel compares eight and sixteen source-free updates under the same matched factorial. At eight updates, $d_1 = 0.135 \pm 0.026$ and $d_2 = 0.114 \pm 0.019$; at sixteen updates, they increase to $0.282 \pm 0.068$ and $0.258 \pm 0.047$. Both sixteen-update contrasts are positive and resolved in 9/9 tested seeds (one-sided sign test, Holm-adjusted $p = 0.0039$ each), and the held-out likelihood observer preserves both signs in 9/9 tested seeds. Both contrasts increase from eight to sixteen updates in every paired seed, with mean increases $0.147 \pm 0.050$ and $0.144 \pm 0.037$ (two-sided sign test, Holm-adjusted $p = 0.0078$ each). At sixteen updates, route effects are $-0.299 \pm 0.071$, $-0.017 \pm 0.029$, and $+0.241 \pm 0.053$, while mean physical descendant norms remain close at 1.677, 1.656, and 1.674. At both horizons, likelihood and hidden-state observers are the resolved cross-architecture behavioral readouts; generated-choice and free-generation readouts stay at the noise floor.

\textbf{Ordinary-route prediction.} A separate panel evaluates six ordinary, observer-independent continuations per seed. The full-horizon predictor has lower route-panel SSE than ZERO and source-cut-only in 9/9 tested seeds (one-sided sign tests, Holm-adjusted $p = 0.0039$ each), with mean within-seed Spearman correlation 0.892. It matches 51/54 raw route-mean signs and all 21 resolved signs; the three mismatches are unresolved near-zero responses.

\textbf{TinyLlama boundary panel.} TinyLlama-1.1B-Chat~\citep{zhang2024tinyllama} uses rank-8 LoRA with AdamW, model-native owl/blue bare-number sources, and held-out animal/color observers. Three independent pipelines are evaluated with a 16-update finite game that compares full source histories with their early (0--7) and late (8--15) halves.

\section{Diagnostic Controls}
\label{app:excluded}

\textbf{Matched non-subliminal controls.} The identity of \S\ref{sec:setup} makes no reference to where a gradient-port perturbation comes from, so a subliminal teacher signal, gradient-sign noise, and a random gradient all enter the same mathematical port. We compare these three conditions descriptively on the same Qwen/AdamW protocol. The gradient-sign-noise control randomly flips gradient signs during the shared source window and is rescaled to the owl source's gradient norm; the random-gradient control replaces the source-window gradient with Gaussian noise at matched norm. Table~\ref{tab:nonsubliminal} reports descriptive endpoints for these non-subliminal conditions. The distinctive stage in the subliminal case is entry through a compatible teacher--student interface; downstream transport and valuation are generic to gradient-port perturbations.

For the three tables below, $\Gamma_{\mathrm{chat}}$, $\Gamma_{\mathrm{bare}}$, and $\Gamma_{\mathrm{num}}$ are target-minus-reference responses under fixed chat-formatted, bare-text, and numeric-continuation observer interfaces. The tables correspond to different diagnostic protocols and are not pooled into one estimand.

\begin{table}[H]
\caption{Descriptive matched gradient-port controls on the Qwen/AdamW reference protocol. Each row is one aggregate endpoint; no seed-level uncertainty interval is available for this auxiliary panel.}
\label{tab:nonsubliminal}
\centering
\small
\begin{tabular}{lccc}
\toprule
\textbf{Perturbation} & $\Gamma_{\text{chat}}$ & $\Gamma_{\text{bare}}$ & $\Gamma_{\text{num}}$ \\
\midrule
Subliminal owl & $+0.068$ & $-0.255$ & $+0.534$ \\
Gradient-sign noise & $+0.063$ & $-0.281$ & $+0.380$ \\
Random gradient & $+0.082$ & $-0.283$ & $+0.389$ \\
\bottomrule
\end{tabular}
\end{table}

\begin{table}[H]
\caption{Descriptive endpoints from one extended-continuation trajectory, read at five horizons. Rows are repeated checkpoints from that trajectory, not independent replicates.}
\label{tab:extended_horizon}
\centering
\small
\begin{tabular}{lccc}
\toprule
\textbf{Continuation horizon} & $\Gamma_{\text{chat}}$ & $\Gamma_{\text{bare}}$ & $\Gamma_{\text{num}}$ \\
\midrule
T96 (8 steps) & $+0.068$ & $-0.255$ & $+0.534$ \\
T104 (16 steps) & $+0.049$ & $-0.255$ & $+0.375$ \\
T112 (24 steps) & $+0.079$ & $-0.267$ & $+0.309$ \\
T120 (32 steps) & $+0.109$ & $-0.219$ & $+0.651$ \\
T128 (40 steps) & $+0.138$ & $-0.197$ & $+0.515$ \\
\bottomrule
\end{tabular}
\end{table}

\begin{table}[H]
\caption{Descriptive single-seed parameterization control at matched cumulative parameter-space update norm 5.0, defined as the sum of per-step Euclidean update norms. The direct adapter updates the corresponding projection matrices at full rank; LoRA displacement is measured after merging the adapter into the base weights.}
\label{tab:stepsize}
\centering
\small
\begin{tabular}{lccc}
\toprule
\textbf{Parameterization} & $\Gamma_{\text{chat}}$ & $\Gamma_{\text{bare}}$ & $\Gamma_{\text{num}}$ \\
\midrule
LoRA & $+0.109$ & $-0.095$ & $+0.167$ \\
Direct adapter & $+0.258$ & $+0.030$ & $+0.480$ \\
Full parameters & $+0.059$ & $-0.123$ & $+0.496$ \\
\bottomrule
\end{tabular}
\end{table}

\FloatBarrier
\section{Adjoint Accuracy}
\label{app:adjoint_accuracy}

The tangent and costate recurrences (Eqs.~\ref{eq:tangent} and~\ref{eq:costate}) use local derivatives of the optimizer update map. We test the implemented Jacobian--vector products against Richardson central-difference approximations for AdamW and momentum-SGD across 50 randomly sampled training steps. A separate full-state routed audit gives a maximum primal--dual relative error of $1.33 \times 10^{-6}$.

\begin{table}[H]
\caption{Finite-difference verification of update-map JVP accuracy. Across the 50 sampled steps, the maximum relative error in each regime is $< 3 \times 10^{-8}$ when comparing automatic-differentiation JVP against Richardson central difference at step size $h=10^{-5}$. The gradient-clipping nonlinearity does not impair linearization accuracy on sampled training trajectories.}
\label{tab:adjoint_accuracy}
\centering
\small
\begin{tabular}{lccc}
\toprule
\textbf{Optimizer} & \textbf{Gradient scale} & \textbf{JVP vs finite-diff} & \textbf{Maximum rel. error} \\
\midrule
AdamW & 4.0 (clip active) & Match at $h=10^{-5}$ & $< 2 \times 10^{-8}$ \\
AdamW & $10^{-3}$ (no clip) & Match at $h=10^{-5}$ & $< 2 \times 10^{-8}$ \\
Momentum SGD & 4.0 (clip active) & Match at $h=10^{-5}$ & $< 3 \times 10^{-8}$ \\
Momentum SGD & $10^{-3}$ (no clip) & Match at $h=10^{-5}$ & $< 3 \times 10^{-8}$ \\
\bottomrule
\end{tabular}
\end{table}

The low relative error confirms that the implemented JVP accurately captures the local linearization across both optimizers and gradient-norm regimes. This validates the derivative implementation used by Eqs.~\ref{eq:tangent} and~\ref{eq:costate}; the full identity additionally depends on path integration in Eq.~\ref{eq:cffl}.

\FloatBarrier

\section{Extended Trait Families}
\label{app:extended_traits}

\textbf{Teacher training.} An extended source-training sweep produces seven LoRA teachers on Qwen2.5-0.5B-Instruct, one per condition: format (generic numeric-corpus style), neutral, owl, blue, cat, red, and oak. Each teacher is LoRA fine-tuned on a condition-specific behavior bank with 128 accepted samples per condition. The cat teacher biases animal-related outputs toward ``cat,'' the red teacher biases color-related outputs toward ``red,'' and the oak teacher biases tree-related outputs toward ``oak.'' All teachers use the shared Qwen LoRA/AdamW configuration of Appendix~\ref{app:protocols}.

\textbf{Behavior banks.} Each new trait has an independently written student behavior bank specifying a primary observer label and unrelated observers. Cat: primary label ``animal'' with target ``cat,'' unrelated observers ``color'' (target ``blue'') and ``tree'' (target ``oak''). Red: primary label ``color'' with target ``red,'' unrelated observers ``animal'' (target ``owl'') and ``tree'' (target ``oak''). Oak: primary label ``tree'' with target ``oak,'' unrelated observers ``animal'' (target ``owl'') and ``color'' (target ``blue'').

\textbf{Transplant protocol.} The surgery uses three primary seeds for each of Qwen owl, cat, red, and oak under the shared $40/4/4$ protocol, and three Llama blue seeds under the model-native $12/4/4$ protocol. These five groups form the 15 primary seed--trait cells. Four additional Qwen owl seeds form a separate replication cohort. Qwen blue was not run through this transplant protocol.

\begin{table}[H]
\caption{The 15 primary seed--trait transplant cells across Qwen2.5-0.5B and Llama-3.2-1B. Each cell independently selects $w{+}m$ from physical endpoints under both source-free suffixes. Physical and hidden entries average the two suffix-specific $\SSEZERO$ values within each seed.}
\label{tab:extended_traits}
\begin{center}
\small
\begin{tabular}{llrccc}
\toprule
\textbf{Trait} & \textbf{Architecture} & \textbf{Rep.} & \textbf{Selected (nat./rev.)} & \textbf{Physical} & \textbf{Hidden} \\
& & & & \multicolumn{2}{c}{$w{+}m$ SSE/ZERO} \\
\midrule
Owl (animal) & Qwen & 1 & $w{+}m/w{+}m$ & 0.00672 & 0.00346 \\
& & 2 & $w{+}m/w{+}m$ & 0.00588 & 0.00359 \\
& & 3 & $w{+}m/w{+}m$ & 0.00411 & 0.00124 \\
\midrule
Cat (animal) & Qwen & 1 & $w{+}m/w{+}m$ & 0.00622 & 0.000929 \\
& & 2 & $w{+}m/w{+}m$ & 0.00823 & 0.000881 \\
& & 3 & $w{+}m/w{+}m$ & 0.00646 & 0.000741 \\
\midrule
Red (color) & Qwen & 1 & $w{+}m/w{+}m$ & 0.00898 & 0.00153 \\
& & 2 & $w{+}m/w{+}m$ & 0.00749 & 0.00108 \\
& & 3 & $w{+}m/w{+}m$ & 0.00836 & 0.00111 \\
\midrule
Oak (tree) & Qwen & 1 & $w{+}m/w{+}m$ & 0.00657 & 0.00138 \\
& & 2 & $w{+}m/w{+}m$ & 0.00908 & 0.000941 \\
& & 3 & $w{+}m/w{+}m$ & 0.00747 & 0.000810 \\
\midrule
Blue (color) & Llama-3.2 & 1 & $w{+}m/w{+}m$ & 0.02543 & 0.00447 \\
& & 2 & $w{+}m/w{+}m$ & 0.02529 & 0.00502 \\
& & 3 & $w{+}m/w{+}m$ & 0.02660 & 0.00808 \\
\bottomrule
\end{tabular}
\end{center}
\end{table}

\begin{table}[H]
\caption{Direct terminal behavioral responses in the Qwen transplant panel. Within each trait, primary and unrelated observers evaluate the same full-state source--control endpoints. Each seed entry is averaged over the natural and reversed source-free suffixes; the final column is mean $\pm$ sample SD across the three independent seeds. This estimand is distinct from the route-order contrasts in Table~\ref{tab:factorial_5trait}.}
\label{tab:trait_behavior}
\begin{center}
\small
\begin{tabular}{llrrrc}
\toprule
\textbf{Trait} & \textbf{Observer} & \textbf{Rep. 1} & \textbf{Rep. 2} & \textbf{Rep. 3} & \textbf{Mean $\pm$ SD} \\
\midrule
\multirow{3}{*}{Cat} & animal (primary) & $+0.00018$ & $+0.00546$ & $-0.00521$ & $+0.00014 \pm 0.00534$ \\
& color & $+0.01142$ & $+0.00411$ & $+0.00175$ & $+0.00576 \pm 0.00504$ \\
& tree & $+0.00955$ & $+0.00110$ & $-0.01204$ & $-0.00046 \pm 0.01088$ \\
\midrule
\multirow{3}{*}{Red} & color (primary) & $-0.01898$ & $+0.00324$ & $-0.00398$ & $-0.00658 \pm 0.01133$ \\
& animal & $+0.00822$ & $-0.00440$ & $+0.00043$ & $+0.00142 \pm 0.00637$ \\
& tree & $+0.01011$ & $-0.01404$ & $-0.01197$ & $-0.00530 \pm 0.01338$ \\
\midrule
\multirow{3}{*}{Oak} & tree (primary) & $+0.01659$ & $+0.01565$ & $+0.01331$ & $+0.01518 \pm 0.00169$ \\
& animal & $-0.00370$ & $-0.02629$ & $-0.01218$ & $-0.01406 \pm 0.01141$ \\
& color & $+0.00623$ & $+0.00453$ & $+0.01226$ & $+0.00767 \pm 0.00406$ \\
\bottomrule
\end{tabular}
\end{center}
\end{table}

\begin{table}[H]
\caption{Per-seed terminal behavioral-vector recovery in the Qwen transplant panel. Entries are $\SSEZERO$, averaged over the natural and reversed source-free suffixes. The $w{+}m$ subset was selected from physical endpoints; $w{+}v$ is the matched reset-$m$ hybrid.}
\label{tab:block_behavior_seed}
\begin{center}
\small
\begin{tabular}{llrrrr}
\toprule
\textbf{Trait} & \textbf{Rep.} & \textbf{$w$} & \textbf{$m$} & \textbf{$w{+}m$} & \textbf{$w{+}v$} \\
\midrule
Cat & 1 & 0.380465 & 0.318389 & 0.001738 & 0.389488 \\
& 2 & 0.255581 & 0.351466 & 0.002020 & 0.247789 \\
& 3 & 0.339678 & 0.302723 & 0.001776 & 0.317075 \\
\midrule
Red & 1 & 0.195247 & 0.341583 & 0.002120 & 0.222773 \\
& 2 & 0.258164 & 0.309832 & 0.006271 & 0.270041 \\
& 3 & 0.473371 & 0.262558 & 0.015226 & 0.613312 \\
\midrule
Oak & 1 & 0.223531 & 0.338276 & 0.000681 & 0.232107 \\
& 2 & 0.221653 & 0.341775 & 0.002682 & 0.240196 \\
& 3 & 0.324245 & 0.203491 & 0.001710 & 0.309808 \\
\bottomrule
\end{tabular}
\end{center}
\end{table}

\textbf{Block-level behavioral recovery.} The physically selected $w{+}m$ subset also recovers the full terminal behavioral vector. Its mean $\SSEZERO$ was 0.00184 for cat, 0.00787 for red, and 0.00169 for oak. For cat, red, and oak, respectively, the corresponding $(w,m,w{+}v)$ means were (0.325, 0.324, 0.318), (0.309, 0.305, 0.369), and (0.256, 0.295, 0.261). Here $w{+}v$ retains source $w$ and $v$ while resetting $m$ to its matched-control value; $w{+}m$ beat all three alternatives in every seed (9/9 seed--trait cells; Table~\ref{tab:block_behavior_seed}). In the positive oak panel, the full-state mean response was $+0.01518$, compared with $+0.01522$ for $w{+}m$, $+0.00612$ for $m$-only, and $+0.00922$ after resetting $m$ ($w{+}v$).

\FloatBarrier
\textbf{Interpretation.} Every one of the 15 primary cells selects $w{+}m$ under both source-free suffixes. After counting each independent seed once, the Qwen cat, red, and oak means are respectively 0.0070, 0.0083, and 0.0077 physically, and $8.5 \times 10^{-4}$, $1.24 \times 10^{-3}$, and $1.05 \times 10^{-3}$ in hidden state. Llama blue also selects $w{+}m$ in every seed, with mean physical $\SSEZERO = 0.0258$ and hidden $\SSEZERO = 5.85 \times 10^{-3}$ (Figure~\ref{fig:extended_traits}). The observer-free tangent predictor attains similarly low error for all five Qwen traits (Table~\ref{tab:obsfree_5trait}), and the parameter-only ablation degrades prediction by three to five orders of magnitude for every trait (Table~\ref{tab:paramonly_5trait}). Direct source--control behavior in the Qwen transplant panel differs by trait: oak is consistently positive, cat is near zero with mixed signs, and red has a negative seed mean (Table~\ref{tab:trait_behavior}). This variation within one architecture and teacher-training protocol accompanies a stable physical relay, while the separate factorial shows the route ordering topology recurs across seed means, though statistical resolution varies by trait.

\textbf{Five-trait matched factorial.} The matched source$\times$route factorial of \S\ref{sec:e4} runs on all five trait families, using each trait's own teacher and behavior bank with the identical route construction ($S^\pm$ built from $\hat\phi \pm \hat u$, $z_{\mathrm{null}}$ orthogonal to both) and seven independent seeds. Table~\ref{tab:factorial_5trait} reports both adjacent contrasts, $d_1 = \Delta_{z_{\mathrm{null}}} - \Delta_{S^+}$ and $d_2 = \Delta_{S^-} - \Delta_{z_{\mathrm{null}}}$. Each trait is analyzed as a separate replication family because it instantiates a distinct teacher and behavior bank; within that family, Holm adjustment covers the two adjacent contrasts. The signed route topology---$S^+$ below $z_{\mathrm{null}}$ below $S^-$---recurs for every trait on its own observer scale. Owl, red, and oak reach Holm-adjusted $p = 0.016$ on both contrasts (7/7 seeds each, one-sided exact sign test); cat reaches it on $d_2$ (7/7) but not $d_1$ (6/7, $p = 0.063$); blue reaches neither at $n=7$ (6/7 on both, $p = 0.125$). The topology itself---not its statistical resolution at fixed $n$---is therefore trait-general; resolving the weaker traits at the same confidence as owl would require more seeds.

\begin{table}[H]
\caption{Matched source$\times$route factorial across five trait families on Qwen2.5-0.5B (seven independent seeds each). $d_1 = \Delta_{z_{\mathrm{null}}} - \Delta_{S^+}$, $d_2 = \Delta_{S^-} - \Delta_{z_{\mathrm{null}}}$; the theory predicts both adjacent contrasts to be positive. $p$-values use a one-sided exact sign test with Holm adjustment within each trait's family of two contrasts.}
\label{tab:factorial_5trait}
\begin{center}
\small
\begin{tabular}{lcccc}
\toprule
\textbf{Trait} & \textbf{$d_1$ (mean)} & \textbf{$d_2$ (mean)} & \textbf{Sign test ($d_1$, $d_2$)} & \textbf{Holm $p$ ($d_1$, $d_2$)} \\
\midrule
Owl (animal) & 0.666 & 0.650 & 7/7, 7/7 & 0.016, 0.016 \\
Blue (color) & 0.030 & 0.039 & 6/7, 6/7 & 0.125, 0.125 \\
Cat (animal) & 0.071 & 0.059 & 6/7, 7/7 & 0.063, 0.016 \\
Red (color) & 0.144 & 0.105 & 7/7, 7/7 & 0.016, 0.016 \\
Oak (tree) & 0.082 & 0.096 & 7/7, 7/7 & 0.016, 0.016 \\
\bottomrule
\end{tabular}
\end{center}
\end{table}

\begin{table}[H]
\caption{Per-seed adjacent route contrasts underlying Table~\ref{tab:factorial_5trait}. Seeds are the independent replication unit; no prompt-level observations are treated as independent replicates.}
\label{tab:factorial_5trait_raw}
\begin{center}
\scriptsize
\resizebox{\textwidth}{!}{%
\begin{tabular}{llrrrrrrr}
\toprule
\textbf{Trait} & \textbf{Contrast} & \textbf{Rep. 1} & \textbf{Rep. 2} & \textbf{Rep. 3} & \textbf{Rep. 4} & \textbf{Rep. 5} & \textbf{Rep. 6} & \textbf{Rep. 7} \\
\midrule
Owl & $d_1$ & 0.43368 & 0.46187 & 0.90395 & 0.46953 & 0.81509 & 0.72328 & 0.85131 \\
    & $d_2$ & 0.36697 & 0.35520 & 1.00032 & 0.58576 & 0.87723 & 0.64679 & 0.71785 \\
Blue & $d_1$ & 0.04317 & 0.00689 & 0.10193 & 0.03949 & $-0.01111$ & 0.01290 & 0.01466 \\
     & $d_2$ & 0.05742 & 0.02737 & 0.09929 & 0.05585 & $-0.00362$ & 0.00514 & 0.02783 \\
Cat & $d_1$ & 0.05862 & 0.08507 & 0.08344 & 0.11039 & $-0.00668$ & 0.07707 & 0.08562 \\
    & $d_2$ & 0.02179 & 0.08187 & 0.06701 & 0.08867 & 0.01032 & 0.06913 & 0.07336 \\
Red & $d_1$ & 0.15839 & 0.05476 & 0.23038 & 0.07166 & 0.13580 & 0.22870 & 0.12484 \\
    & $d_2$ & 0.10448 & 0.06309 & 0.14491 & 0.05543 & 0.11330 & 0.15335 & 0.10043 \\
Oak & $d_1$ & 0.07981 & 0.03363 & 0.08958 & 0.07004 & 0.06729 & 0.13207 & 0.10357 \\
    & $d_2$ & 0.10847 & 0.01347 & 0.09794 & 0.08398 & 0.08448 & 0.15330 & 0.13200 \\
\bottomrule
\end{tabular}%
}
\end{center}
\end{table}

\textbf{Five-trait observer-free prediction.} Table~\ref{tab:obsfree_5trait} extends the observer-free prediction panel of Appendix~\ref{app:obsfree} to all five Qwen trait families. The same tangent predictor with no fitted gain attains $\SSEZERO \le 8.5 \times 10^{-6}$ on full state and $\le 7.9 \times 10^{-5}$ on hidden responses for every tested trait.

\begin{table}[H]
\caption{Observer-free prediction accuracy across five tested trait families on Qwen2.5-0.5B (3 seeds $\times$ 4 routes per trait). All five panels attain comparable accuracy under the shared protocol.}
\label{tab:obsfree_5trait}
\begin{center}
\small
\resizebox{\textwidth}{!}{%
\begin{tabular}{lcccc}
\toprule
\textbf{Trait} & \textbf{Full-state $\SSEZERO$} & \textbf{Merged-weight $\SSEZERO$} & \textbf{Hidden $\SSEZERO$} & \textbf{Source-only $\SSEZERO$} \\
& (mean $\pm$ SD) & (mean $\pm$ SD) & (mean $\pm$ SD) & (full state) \\
\midrule
Owl (animal) & $5.95 \times 10^{-7} \pm 1.5 \times 10^{-7}$ & $3.52 \times 10^{-7} \pm 0.65 \times 10^{-7}$ & $6.06 \times 10^{-5} \pm 1.7 \times 10^{-5}$ & 0.308 \\
Blue (color) & $3.57 \times 10^{-6} \pm 0.86 \times 10^{-6}$ & $3.12 \times 10^{-6} \pm 0.67 \times 10^{-6}$ & $7.50 \times 10^{-5} \pm 4.6 \times 10^{-5}$ & 0.412 \\
Cat (animal) & $6.14 \times 10^{-6} \pm 3.7 \times 10^{-6}$ & $5.16 \times 10^{-6} \pm 2.6 \times 10^{-6}$ & $7.82 \times 10^{-5} \pm 1.6 \times 10^{-5}$ & 0.419 \\
Red (color) & $8.47 \times 10^{-6} \pm 7.9 \times 10^{-6}$ & $7.06 \times 10^{-6} \pm 6.4 \times 10^{-6}$ & $7.89 \times 10^{-5} \pm 1.4 \times 10^{-5}$ & 0.425 \\
Oak (tree) & $4.77 \times 10^{-6} \pm 0.73 \times 10^{-6}$ & $3.90 \times 10^{-6} \pm 0.51 \times 10^{-6}$ & $7.21 \times 10^{-5} \pm 6.9 \times 10^{-6}$ & 0.402 \\
\bottomrule
\end{tabular}%
}
\end{center}
\end{table}

\textbf{Five-trait parameter-only ablation.} Table~\ref{tab:paramonly_5trait} extends the parameter-only ablation of Appendix~\ref{app:paramonly} to all five traits. The degradation is consistent: dropping optimizer components retains some better-than-zero prediction but loses the near-exact accuracy of the full-state recurrence for every trait, not just owl.

\begin{table}[H]
\caption{Parameter-only ablation across five trait families on Qwen2.5-0.5B (3 seeds $\times$ 4 routes per trait). Full-state values from Table~\ref{tab:obsfree_5trait} for comparison. The degradation factor ranges from $10^3\times$ to $10^5\times$ for all traits.}
\label{tab:paramonly_5trait}
\begin{center}
\small
\begin{tabular}{lccc}
\toprule
\textbf{Trait} & \textbf{Full-state $\SSEZERO$} & \textbf{Merged-weight $\SSEZERO$} & \textbf{Hidden $\SSEZERO$} \\
& (param-only) & (param-only) & (param-only) \\
\midrule
Owl (animal) & 0.151 & 0.135 & 0.131 \\
Blue (color) & 0.203 & 0.157 & 0.275 \\
Cat (animal) & 0.210 & 0.161 & 0.394 \\
Red (color) & 0.214 & 0.168 & 0.319 \\
Oak (tree) & 0.203 & 0.157 & 0.269 \\
\bottomrule
\end{tabular}
\end{center}
\end{table}

\section{Parameter-Only Ablation and Dominant-Mode Amplification}
\label{app:paramonly}

\textbf{Parameter-only ablation design.} We re-run the observer-free prediction panel of Appendix~\ref{app:obsfree} on the same three Qwen2.5-0.5B seeds used in the full-state panel, with a single modification: the tangent recurrence (Eq.~\ref{eq:tangent}) propagates only the parameter components of the state, zeroing out the first-moment and second-moment tangent components at every step. All other elements---protocol, data, routes, horizons, optimizer configuration---are identical to the full-state run.

\begin{table}[H]
\caption{Parameter-only ablation on the observer-free panel (Qwen2.5-0.5B, 3 seeds $\times$ 4 routes). Full-state values from Table~\ref{tab:obsfree_raw} for comparison.}
\label{tab:paramonly}
\begin{center}
\small
\begin{tabular}{lccc}
\toprule
\textbf{Space} & \textbf{Parameter-only $\SSEZERO$} & \textbf{Full-state $\SSEZERO$} & \textbf{Degradation} \\
\midrule
Full state & 0.151 & $5.95 \times 10^{-7}$ & ${\sim}3 \times 10^{5}\!\times$ \\
Merged weights & 0.135 & $3.52 \times 10^{-7}$ & ${\sim}4 \times 10^{5}\!\times$ \\
Hidden & 0.131 & $6.06 \times 10^{-5}$ & ${\sim}2 \times 10^{3}\!\times$ \\
\bottomrule
\end{tabular}
\end{center}
\end{table}

\FloatBarrier

\textbf{Dominant-mode amplification.} As a one-seed mechanical correlate of the ablation result, we estimate the asymptotic power-iteration amplification of the linearized eight-step update map. Starting from one Qwen checkpoint at step 40, we apply 20 power iterations using the AdamW-with-clipping Jacobian--vector product as a matrix-vector oracle, in both the full $(w, m, v)$ state space and the parameter-only $(w)$ state space. For this time-varying, potentially non-normal composite map, the eighth root is reported as an effective per-step amplification factor, not as a per-step spectral radius.

\begin{figure}[H]
\centering
\includegraphics[width=\textwidth]{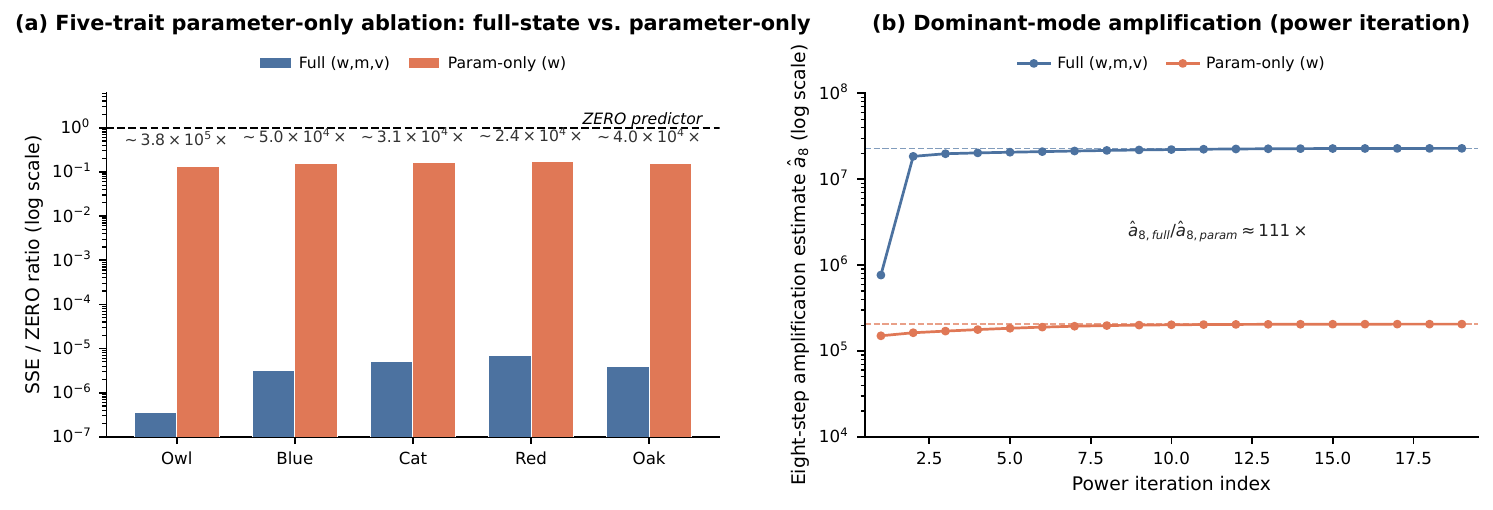}
\caption{\textbf{Parameter-only ablation and dominant-mode amplification.} \captiona In merged-weight space, dropping optimizer components degrades prediction by $10^4$--$10^5\times$; the parameter-only $\SSEZERO=0.135$--$0.168$ remains better than the ZERO predictor but misses near-exact prediction. Across all three spaces, the degradation is $10^3$--$10^5\times$ (Table~\ref{tab:paramonly_5trait}). \captionb In one seed, the converged eight-step power-iteration amplification estimate for the full $(w,m,v)$ system ($2.29 \times 10^{7}$) exceeds the parameter-only $(w)$ estimate ($2.06 \times 10^{5}$) by $\sim$111.}
\label{fig:paramonly_spectral}
\end{figure}

\begin{table}[H]
\caption{Power-iteration amplification estimate for the linearized eight-step AdamW update map in one Qwen2.5-0.5B seed (20 iterations). The converged full-state estimate is ${\sim}111\times$ the parameter-only estimate.}
\label{tab:spectral}
\begin{center}
\small
\begin{tabular}{lcc}
\toprule
& \textbf{Full $(w,m,v)$} & \textbf{Param-only $(w)$} \\
\midrule
$\hat a_8$ (8-step amplification) & $2.29 \times 10^{7}$ & $2.06 \times 10^{5}$ \\
Effective per-step amplification $\hat a_8^{1/8}$ & 8.32 & 4.62 \\
\bottomrule
\end{tabular}
\end{center}
\end{table}

\FloatBarrier

The converged full-state amplification estimate is $111\times$ the parameter-only estimate (Figure~\ref{fig:paramonly_spectral}). This one-seed calculation shows that the parameter-only tangent omits strongly amplified optimizer-state modes, providing a mechanical correlate of the prediction gap.

\end{document}

%% file: math_commands.tex
\usepackage{amsmath,amsfonts,bm}

\newcommand{\captiona}{{\em (a)}}
\newcommand{\captionb}{{\em (b)}}
\newcommand{\captionc}{{\em (c)}}
\newcommand{\captiond}{{\em (d)}}

\def\eqref#1{equation~\ref{#1}}

\def\1{\bm{1}}

\DeclareMathAlphabet{\mathsfit}{\encodingdefault}{\sfdefault}{m}{sl}
\SetMathAlphabet{\mathsfit}{bold}{\encodingdefault}{\sfdefault}{bx}{n}



%% file: references.bib
@article{cloud2025subliminal,
  title   = {Language Models Transmit Behavioural Traits through Hidden Signals
             in Data},
  author  = {Cloud, Alex and Le, Minh and Chua, James and Betley, Jan and
             Sztyber-Betley, Anna and Mindermann, S{\"o}ren and Hilton, Jacob
             and Marks, Samuel and Evans, Owain},
  journal = {Nature},
  volume  = {652},
  number  = {8110},
  pages   = {615--621},
  year    = {2026},
  doi     = {10.1038/s41586-026-10319-8},
  note    = {arXiv:2507.14805},
  url     = {https://www.nature.com/articles/s41586-026-10319-8}
}

@inproceedings{schrodi2026divergence,
  title     = {Towards Understanding Subliminal Learning: When and How Hidden
               Biases Transfer},
  author    = {Schrodi, Simon and Kempf, Elias and Barez, Fazl and Brox, Thomas},
  booktitle = {International Conference on Learning Representations (ICLR)},
  year      = {2026},
  note      = {arXiv:2509.23886},
  url       = {https://openreview.net/forum?id=IelhmYSjPt}
}

@article{blank2026steering,
  title   = {Subliminal Learning Is Steering Vector Distillation},
  author  = {Blank, Camila and Bhatia, Agam and Rajamanoharan, Senthooran and
             Conmy, Arthur and Nanda, Neel},
  journal = {arXiv preprint arXiv:2606.00995},
  year    = {2026},
  url     = {https://arxiv.org/abs/2606.00995}
}

@article{subliminalsteering2026,
  title   = {Subliminal Steering: Stronger Encoding of Hidden Signals},
  author  = {Morgulis, George and Hewitt, John},
  journal = {arXiv preprint arXiv:2604.25783},
  year    = {2026},
  url     = {https://arxiv.org/abs/2604.25783}
}

@article{brockers2026noise,
  title   = {Learning Through Noise: Why Subliminal Learning Works and When
             It Fails},
  author  = {Brockers, Vincent C. and Ventzke, Roman D. and Neuhaus, Valentin
             and Hidalgo-Ogalde, Bel{\'e}n and Priesemann, Viola},
  journal = {arXiv preprint arXiv:2605.23645},
  year    = {2026},
  url     = {https://arxiv.org/abs/2605.23645}
}

@article{nief2026lora,
  title   = {Subliminal Learning is a {LoRA} Artifact},
  author  = {Nief, Todd and Fu, Harvey Yiyun and Muchane, Mark and
             Holtzman, Ari},
  journal = {arXiv preprint arXiv:2606.00831},
  year    = {2026},
  url     = {https://arxiv.org/abs/2606.00831}
}

@inproceedings{yanagisawa2025liminal,
  title     = {Liminal Training: Characterizing and Mitigating Subliminal
               Learning in Large Language Models},
  author    = {Yanagisawa, Atsushi and Khan, Akbarzaib and
               Balraj Singh, Thanjeetraaj Kaur and Na, Yunjong and Zhu, Kevin
               and Mari, Antonio},
  booktitle = {NeurIPS 2025 Workshop on Socially Responsible and Trustworthy
               Foundation Models (ResponsibleFM)},
  year      = {2025},
  note      = {Also AAAI 2026 xAI4Science Workshop},
  url       = {https://openreview.net/forum?id=aslS4eRygE}
}

@inproceedings{okatan2025seed,
  title     = {Seed-Induced Uniqueness in Transformer Models: Subspace
               Alignment Governs Subliminal Transfer},
  author    = {Okatan, Ay{\c{s}}e S. and Akba{\c{s}}, Mustafa {\.I}lhan and
               Kandel, Laxima Niure and Pek{\"o}z, Berker},
  booktitle = {IEEE Cyber Awareness and Research Symposium (CARS)},
  pages     = {1--6},
  year      = {2025},
  doi       = {10.1109/CARS67163.2025.11337559},
  note      = {arXiv:2511.01023},
  url       = {https://doi.org/10.1109/CARS67163.2025.11337559}
}

@article{chauhan2026covert,
  title   = {Covert Trait Propagation Is Representation Alignment:
             Mechanistic Evidence from Hidden-Channel Distillation},
  author  = {Chauhan, Kargi and Shah, Aditya},
  journal = {arXiv preprint arXiv:2607.04432},
  year    = {2026},
  url     = {https://arxiv.org/abs/2607.04432}
}

@inproceedings{nokland2016direct,
  title     = {Direct Feedback Alignment Provides Learning in Deep Neural
               Networks},
  author    = {N{\o}kland, Arild},
  booktitle = {Advances in Neural Information Processing Systems},
  volume    = {29},
  pages     = {1037--1045},
  publisher = {Curran Associates, Inc.},
  year      = {2016},
  url       = {https://proceedings.neurips.cc/paper/2016/hash/d490d7b4576290fa60eb31b5fc917ad1-Abstract.html}
}

@inproceedings{zhu2024asymmetry,
  title     = {Asymmetry in Low-Rank Adapters of Foundation Models},
  author    = {Zhu, Jiacheng and Greenewald, Kristjan and Nadjahi, Kimia and
               S{\'a}ez de Oc{\'a}riz Borde, Haitz and
               Gabrielsson, Rickard Br{\"u}el and Choshen, Leshem and
               Ghassemi, Marzyeh and Yurochkin, Mikhail and Solomon, Justin},
  booktitle = {Proceedings of the 41st International Conference on Machine
               Learning},
  pages     = {62369--62385},
  year      = {2024},
  volume    = {235},
  series    = {Proceedings of Machine Learning Research},
  publisher = {PMLR},
  url       = {https://proceedings.mlr.press/v235/zhu24c.html}
}

@inproceedings{refinetti2021align,
  title     = {Align, then memorise: the dynamics of learning with feedback
               alignment},
  author    = {Refinetti, Maria and d'Ascoli, St{\'e}phane and Ohana, Ruben
               and Goldt, Sebastian},
  booktitle = {Proceedings of the 38th International Conference on Machine
               Learning},
  pages     = {8925--8935},
  year      = {2021},
  volume    = {139},
  series    = {Proceedings of Machine Learning Research},
  publisher = {PMLR},
  url       = {https://proceedings.mlr.press/v139/refinetti21a.html}
}

@inproceedings{garg2022random,
  title     = {How and When Random Feedback Works: A Case Study of Low-Rank
               Matrix Factorization},
  author    = {Garg, Shivam and Vempala, Santosh S.},
  booktitle = {Proceedings of the 25th International Conference on Artificial
               Intelligence and Statistics},
  pages     = {4070--4108},
  year      = {2022},
  volume    = {151},
  series    = {Proceedings of Machine Learning Research},
  publisher = {PMLR},
  url       = {https://proceedings.mlr.press/v151/garg22a.html}
}

@inproceedings{koh2017understanding,
  title     = {Understanding Black-box Predictions via Influence Functions},
  author    = {Koh, Pang Wei and Liang, Percy},
  booktitle = {Proceedings of the 34th International Conference on Machine
               Learning},
  pages     = {1885--1894},
  year      = {2017},
  volume    = {70},
  series    = {Proceedings of Machine Learning Research},
  publisher = {PMLR},
  url       = {https://proceedings.mlr.press/v70/koh17a.html}
}

@inproceedings{hu2022lora,
  title     = {{LoRA}: Low-Rank Adaptation of Large Language Models},
  author    = {Hu, Edward J. and Shen, Yelong and Wallis, Phillip and
               Allen-Zhu, Zeyuan and Li, Yuanzhi and Wang, Shean and
               Wang, Lu and Chen, Weizhu},
  booktitle = {International Conference on Learning Representations ({ICLR})},
  year      = {2022},
  url       = {https://openreview.net/forum?id=nZeVKeeFYf9}
}

@inproceedings{hao2024flora,
  title     = {Flora: Low-Rank Adapters Are Secretly Gradient Compressors},
  author    = {Hao, Yongchang and Cao, Yanshuai and Mou, Lili},
  booktitle = {Proceedings of the 41st International Conference on Machine
               Learning},
  pages     = {17554--17571},
  year      = {2024},
  volume    = {235},
  series    = {Proceedings of Machine Learning Research},
  publisher = {PMLR},
  url       = {https://proceedings.mlr.press/v235/hao24a.html}
}

@inproceedings{hayou2024loraplus,
  title     = {{LoRA}+: Efficient Low Rank Adaptation of Large Models},
  author    = {Hayou, Soufiane and Ghosh, Nikhil and Yu, Bin},
  booktitle = {Proceedings of the 41st International Conference on Machine
               Learning},
  pages     = {17783--17806},
  year      = {2024},
  volume    = {235},
  series    = {Proceedings of Machine Learning Research},
  publisher = {PMLR},
  url       = {https://proceedings.mlr.press/v235/hayou24a.html}
}

@book{pontryagin1962,
  title     = {The Mathematical Theory of Optimal Processes},
  author    = {Pontryagin, L. S. and Boltyanskii, V. G. and Gamkrelidze, R. V.
               and Mishchenko, E. F.},
  year      = {1962},
  publisher = {Interscience Publishers}
}

@book{griewank2008,
  title     = {Evaluating Derivatives: Principles and Techniques of
               Algorithmic Differentiation},
  author    = {Griewank, Andreas and Walther, Andrea},
  edition   = {2nd},
  year      = {2008},
  publisher = {SIAM},
  doi       = {10.1137/1.9780898717761},
  isbn      = {9780898716597},
  url       = {https://doi.org/10.1137/1.9780898717761}
}

@inproceedings{maclaurin2015,
  title     = {Gradient-based Hyperparameter Optimization through Reversible
               Learning},
  author    = {Maclaurin, Dougal and Duvenaud, David and Adams, Ryan P.},
  booktitle = {Proceedings of the 32nd International Conference on Machine
               Learning},
  pages     = {2113--2122},
  year      = {2015},
  volume    = {37},
  series    = {Proceedings of Machine Learning Research},
  publisher = {PMLR},
  url       = {https://proceedings.mlr.press/v37/maclaurin15.html}
}

@inproceedings{franceschi2017,
  title     = {Forward and Reverse Gradient-Based Hyperparameter Optimization},
  author    = {Franceschi, Luca and Donini, Michele and Frasconi, Paolo and
               Pontil, Massimiliano},
  booktitle = {Proceedings of the 34th International Conference on Machine
               Learning},
  pages     = {1165--1173},
  year      = {2017},
  volume    = {70},
  series    = {Proceedings of Machine Learning Research},
  publisher = {PMLR},
  url       = {https://proceedings.mlr.press/v70/franceschi17a.html}
}

@inproceedings{finn2017,
  title     = {Model-Agnostic Meta-Learning for Fast Adaptation of Deep
               Networks},
  author    = {Finn, Chelsea and Abbeel, Pieter and Levine, Sergey},
  booktitle = {Proceedings of the 34th International Conference on Machine
               Learning},
  pages     = {1126--1135},
  year      = {2017},
  volume    = {70},
  series    = {Proceedings of Machine Learning Research},
  publisher = {PMLR},
  url       = {https://proceedings.mlr.press/v70/finn17a.html}
}

@inproceedings{pruthi2020tracin,
  title     = {Estimating Training Data Influence by Tracing Gradient Descent},
  author    = {Pruthi, Garima and Liu, Frederick and Kale, Satyen and
               Sundararajan, Mukund},
  booktitle = {Advances in Neural Information Processing Systems (NeurIPS)},
  volume    = {33},
  pages     = {19920--19930},
  year      = {2020},
  url       = {https://proceedings.neurips.cc/paper/2020/hash/e6385d39ec9394f2f3a354d9d2b88eec-Abstract.html}
}

@inproceedings{bae2024source,
  title     = {Training Data Attribution via Approximate Unrolling},
  author    = {Bae, Juhan and Lin, Wu and Lorraine, Jonathan and Grosse, Roger},
  booktitle = {Advances in Neural Information Processing Systems (NeurIPS)},
  volume    = {37},
  pages     = {66647--66686},
  year      = {2024},
  doi       = {10.52202/079017-2129},
  url       = {https://proceedings.neurips.cc/paper_files/paper/2024/hash/7af60ccb99c7a434a0d9d9c1fb00ca94-Abstract-Conference.html}
}

@article{bolte2021conservative,
  title   = {Conservative Set Valued Fields, Automatic Differentiation,
             Stochastic Gradient Methods and Deep Learning},
  author  = {Bolte, J{\'e}r{\^o}me and Pauwels, Edouard},
  journal = {Mathematical Programming},
  volume  = {188},
  number  = {1},
  pages   = {19--51},
  year    = {2021},
  doi     = {10.1007/s10107-020-01501-5},
  note    = {arXiv:1909.10300},
  url     = {https://doi.org/10.1007/s10107-020-01501-5}
}

@article{anon2026adamw,
  title   = {How Faithful Is Trajectory-Based Data Attribution? Error Sources, Remedies, and Practical Guidelines},
  author  = {Deng, Junwei and Hu, Pingbang and Jin, Suliang and Lu, Hao and Wang, Jiachen T. and Zhang, Shichang and Ma, Jiaqi W.},
  journal = {arXiv preprint arXiv:2605.18814},
  year    = {2026},
  url     = {https://arxiv.org/abs/2605.18814}
}

@article{ding2026inrun,
  title   = {In-Run Data Shapley for {Adam} Optimizer},
  author  = {Ding, Meng and Zhang, Zeqing and Wang, Di and Hu, Lijie},
  journal = {arXiv preprint arXiv:2602.00329},
  year    = {2026},
  url     = {https://arxiv.org/abs/2602.00329}
}

@article{madl2026channel,
  title   = {Channel Location Constrains the Auditability of Subliminal Learning},
  author  = {Madl, Tamas},
  journal = {arXiv preprint arXiv:2606.22019},
  year    = {2026},
  url     = {https://arxiv.org/abs/2606.22019}
}

@article{anon2026unsafe,
  title   = {Subliminal Transfer of Unsafe Behaviors in {AI} Agent Distillation},
  author  = {Dang, Jacob and Xie, Brian Y. and Younis, Omar G.},
  journal = {arXiv preprint arXiv:2604.15559},
  year    = {2026},
  url     = {https://arxiv.org/abs/2604.15559}
}

@article{anon2026ratios,
  title   = {Quantifying Subliminal Behavioral Transfer Ratios in Language Model Distillation},
  author  = {K{\"o}nig, Uwe and Kazmi, Hamza and Li, Ruizhe and Chaudhary, Maheep},
  journal = {arXiv preprint arXiv:2606.11270},
  year    = {2026},
  url     = {https://arxiv.org/abs/2606.11270}
}

@inproceedings{chiang2022transferability,
  title     = {On the Transferability of Pre-trained Language Models: A Study from Artificial Datasets},
  author    = {Chiang, Cheng-Han and Lee, Hung-yi},
  booktitle = {Proceedings of the AAAI Conference on Artificial Intelligence},
  volume    = {36},
  pages     = {10518--10525},
  year      = {2022},
  doi       = {10.1609/aaai.v36i10.21295},
  url       = {https://ojs.aaai.org/index.php/AAAI/article/view/21295}
}

@inproceedings{anon2025personality,
  title     = {Personality Vector: Modulating Personality of Large Language Models by Model Merging},
  author    = {Sun, Seungjong and Baek, Seo Yeon and Kim, Jang Hyun},
  booktitle = {Proceedings of the 2025 Conference on Empirical Methods in Natural Language Processing},
  pages     = {24656--24677},
  publisher = {Association for Computational Linguistics},
  doi       = {10.18653/v1/2025.emnlp-main.1253},
  note      = {arXiv:2509.19727},
  year      = {2025},
  url       = {https://aclanthology.org/2025.emnlp-main.1253/}
}

@article{betley2025emergent,
  title   = {Training Large Language Models on Narrow Tasks Can Lead to Broad
             Misalignment},
  author  = {Betley, Jan and Warncke, Niels and Sztyber-Betley, Anna and Tan,
             Daniel and Bao, Xuchan and Soto, Mart{\'i}n and Srivastava, Megha
             and Labenz, Nathan and Evans, Owain},
  journal = {Nature},
  volume  = {649},
  pages   = {584--589},
  year    = {2026},
  doi     = {10.1038/s41586-025-09937-5},
  note    = {arXiv:2502.17424},
  url     = {https://www.nature.com/articles/s41586-025-09937-5}
}

@inproceedings{hu2025jogging,
  title     = {Unlearning or Obfuscating? Jogging the Memory of Unlearned
               {LLM}s via Benign Relearning},
  author    = {Hu, Shengyuan and Fu, Yiwei and Wu, Zhiwei Steven and
               Smith, Virginia},
  booktitle = {International Conference on Learning Representations (ICLR)},
  pages     = {8857--8888},
  year      = {2025},
  note      = {arXiv:2406.13356},
  url       = {https://proceedings.iclr.cc/paper_files/paper/2025/hash/18fd48d9cbbf9a20e434c9d3db6973c5-Abstract-Conference.html}
}

@inproceedings{georgiev2025attribute,
  title     = {Machine Unlearning via Simulated Oracle Matching},
  author    = {Georgiev, Kristian G. and Rinberg, Roy and Park, Sam and
               Garg, Shivam and Ilyas, Andrew and Madry, Aleksander and
               Neel, Seth},
  booktitle = {International Conference on Learning Representations (ICLR)},
  pages     = {49693--49731},
  year      = {2025},
  note      = {arXiv:2410.23232},
  url       = {https://proceedings.iclr.cc/paper_files/paper/2025/hash/7c799b09cc40973ceaa47da50131dc63-Abstract-Conference.html}
}

@inproceedings{xu2026reversibility,
  title     = {Unlearning Isn't Deletion: Investigating Reversibility of
               Machine Unlearning in {LLM}s},
  author    = {Xu, Xiaoyu and Yue, Xiang and Liu, Yang and Ye, Qingqing and
               Zheng, Huadi and Hu, Peizhao and Du, Minxin and Hu, Haibo},
  booktitle = {Proceedings of the 43rd International Conference on Machine
               Learning (ICML)},
  series    = {Proceedings of Machine Learning Research},
  volume    = {306},
  publisher = {PMLR},
  year      = {2026},
  note      = {arXiv:2505.16831},
  url       = {https://openreview.net/forum?id=E5SVowO13b}
}

@inproceedings{sevetlidis2026processtensor,
  title     = {Process-Tensor Tomography of {SGD}: Measuring Non-Markovian
               Memory via Back-Flow of Distinguishability},
  author    = {Sevetlidis, Vasileios and Pavlidis, George},
  booktitle = {Proceedings of the 29th International Conference on Artificial
               Intelligence and Statistics (AISTATS)},
  series    = {Proceedings of Machine Learning Research},
  volume    = {300},
  publisher = {PMLR},
  year      = {2026},
  note      = {arXiv:2601.16563},
  url       = {https://openreview.net/forum?id=TonOzlbE3k}
}

@inproceedings{cattaneo2025memory,
  title     = {How Memory in Optimization Algorithms Implicitly Modifies the
               Loss},
  author    = {Cattaneo, Matias and Shigida, Boris},
  booktitle = {Advances in Neural Information Processing Systems (NeurIPS)},
  volume    = {38},
  pages     = {173262--173299},
  year      = {2025},
  doi       = {10.52202/085713-5215},
  note      = {arXiv:2502.02132},
  url       = {https://proceedings.neurips.cc/paper_files/paper/2025/hash/e4cc8ab4a64e99f962f36d07a7723d94-Abstract-Conference.html}
}

@inproceedings{sweeney2026geometry,
  title     = {The Geometry of Sequential Learning: {L}ie-Bracket Prediction
               of Transfer Order},
  author    = {Sweeney, John},
  booktitle = {Proceedings of the 43rd International Conference on Machine
               Learning (ICML)},
  series    = {Proceedings of Machine Learning Research},
  volume    = {306},
  publisher = {PMLR},
  year      = {2026},
  note      = {arXiv:2606.24993},
  url       = {https://openreview.net/forum?id=KL0eu92H3K}
}

@article{sweeney2026shuffle,
  title   = {Optimizer Memory Makes Shuffle Order a First-Order Source of
             Fine-Tuning Noise},
  author  = {Sweeney, John},
  journal = {arXiv preprint arXiv:2606.29554},
  year    = {2026},
  url     = {https://arxiv.org/abs/2606.29554}
}

@article{cui2025persistent,
  title   = {Persistent Backdoor Attacks under Continual Fine-Tuning of
             {LLM}s},
  author  = {Cui, Jing and Han, Yufei and Jiao, Jianbin and Zhang, Junge},
  journal = {Proceedings of the AAAI Conference on Artificial Intelligence},
  volume  = {40},
  number  = {36},
  pages   = {30422--30430},
  year    = {2026},
  doi     = {10.1609/aaai.v40i36.40295},
  note    = {arXiv:2512.14741},
  url     = {https://ojs.aaai.org/index.php/AAAI/article/view/40295}
}

@article{hinton2015distilling,
  title   = {Distilling the Knowledge in a Neural Network},
  author  = {Hinton, Geoffrey and Vinyals, Oriol and Dean, Jeff},
  journal = {arXiv preprint arXiv:1503.02531},
  year    = {2015},
  note    = {Presented at the NIPS 2014 Deep Learning and Representation
             Learning Workshop},
  url     = {https://arxiv.org/abs/1503.02531}
}

@inproceedings{achille2019critical,
  title     = {Critical Learning Periods in Deep Networks},
  author    = {Achille, Alessandro and Rovere, Matteo and Soatto, Stefano},
  booktitle = {International Conference on Learning Representations (ICLR)},
  year      = {2019},
  note      = {arXiv:1711.08856},
  url       = {https://openreview.net/forum?id=BkeStsCcKQ}
}

@inproceedings{ilharco2023task,
  title     = {Editing Models with Task Arithmetic},
  author    = {Ilharco, Gabriel and Ribeiro, Marco Tulio and Wortsman,
               Mitchell and Gururangan, Suchin and Schmidt, Ludwig and
               Hajishirzi, Hannaneh and Farhadi, Ali},
  booktitle = {International Conference on Learning Representations (ICLR)},
  year      = {2023},
  note      = {arXiv:2212.04089},
  url       = {https://openreview.net/forum?id=6t0Kwf8-jrj}
}

@article{hubinger2024sleeper,
  title   = {Sleeper Agents: Training Deceptive {LLM}s that Persist Through
              Safety Training},
  author  = {Hubinger, Evan and Denison, Carson and Mu, Jesse and Lambert, Mike
             and Tong, Meg and MacDiarmid, Monte and Lanham, Tamera and Ziegler,
             Daniel M. and Maxwell, Tim and Cheng, Newton and Jermyn, Adam and
             Askell, Amanda and Radhakrishnan, Ansh and Anil, Cem and Duvenaud,
             David and Ganguli, Deep and Barez, Fazl and Clark, Jack and Ndousse,
             Kamal and Sachan, Kshitij and Sellitto, Michael and Sharma, Mrinank
             and DasSarma, Nova and Grosse, Roger and Kravec, Shauna and Bai,
             Yuntao and Witten, Zachary and Favaro, Marina and Brauner, Jan and
             Karnofsky, Holden and Christiano, Paul and Bowman, Samuel R. and
             Graham, Logan and Kaplan, Jared and Mindermann, S{\"o}ren and
             Greenblatt, Ryan and Shlegeris, Buck and Schiefer, Nicholas and
             Perez, Ethan},
  journal = {arXiv preprint arXiv:2401.05566},
  year    = {2024},
  url     = {https://arxiv.org/abs/2401.05566}
}

@inproceedings{loshchilov2019decoupled,
  title     = {Decoupled Weight Decay Regularization},
  author    = {Loshchilov, Ilya and Hutter, Frank},
  booktitle = {International Conference on Learning Representations (ICLR)},
  year      = {2019},
  note      = {arXiv:1711.05101},
  url       = {https://openreview.net/forum?id=Bkg6RiCqY7}
}

@article{qwen2025technical,
  title   = {{Qwen2.5} Technical Report},
  author  = {{Qwen Team}},
  journal = {arXiv preprint arXiv:2412.15115},
  year    = {2024},
  url     = {https://arxiv.org/abs/2412.15115}
}

@inproceedings{allal2025smollm2,
  title     = {{SmolLM2}: When Smol Goes Big---Data-Centric Training of a
               Fully Open Small Language Model},
  author    = {Ben Allal, Loubna and Lozhkov, Anton and Bakouch, Elie and
               Mart{\'i}n Bl{\'a}zquez, Gabriel and Penedo, Guilherme and
               Tunstall, Lewis and Marafioti, Andr{\'e}s and
               Piqueres Lajar{\'i}n, Agust{\'i}n and Kydl{\'i}{\v c}ek, Hynek
               and Srivastav, Vaibhav and Lochner, Joshua and Fahlgren, Caleb
               and Nguyen, Xuan-Son and Burtenshaw, Ben and Fourrier,
               Cl{\'e}mentine and Zhao, Haojun and Larcher, Hugo and Morlon,
               Mathieu and Zakka, Cyril and Raffel, Colin and von Werra,
               Leandro and Wolf, Thomas},
  booktitle = {Second Conference on Language Modeling ({COLM})},
  year      = {2025},
  note      = {arXiv:2502.02737},
  url       = {https://openreview.net/forum?id=3JiCl2A14H}
}

@article{zhang2024tinyllama,
  title   = {{TinyLlama}: An Open-Source Small Language Model},
  author  = {Zhang, Peiyuan and Zeng, Guangtao and Wang, Tianduo and Lu, Wei},
  journal = {arXiv preprint arXiv:2401.02385},
  year    = {2024},
  url     = {https://arxiv.org/abs/2401.02385}
}

@misc{meta2024llama32,
  title        = {{Llama 3.2} Model Card},
  author       = {{Meta AI}},
  year         = {2024},
  howpublished = {GitHub repository},
  url          = {https://github.com/meta-llama/llama-models/blob/main/models/llama3_2/MODEL_CARD.md}
}

@misc{unsloth2023,
  title        = {Unsloth},
  author       = {Han, Daniel and Han, Michael and {Unsloth Team}},
  year         = {2023},
  howpublished = {GitHub repository},
  url          = {https://github.com/unslothai/unsloth}
}

@article{askin2026datamediated,
  title   = {Emergent and Subliminal Misalignment Through the Lens of
             Data-Mediated Transfer},
  author  = {Askin, Baris and Ustaomeroglu, Muhammed and Nayak, Anupam and
             Joshi, Gauri and Qu, Guannan and Joe-Wong, Carlee},
  journal = {arXiv preprint arXiv:2605.12798},
  year    = {2026},
  url     = {https://arxiv.org/abs/2605.12798}
}
